\documentclass[final,3p,times,authoryear]{elsarticle}

\usepackage{amssymb}
\usepackage{amsmath}

\usepackage[utf8]{inputenc}
\usepackage[T1]{fontenc}
\usepackage{url}
\usepackage{hyperref}
\hypersetup{colorlinks=true, linkcolor=blue, citecolor=blue, urlcolor=blue} 

\usepackage{booktabs}
\usepackage{longtable} 
\usepackage{tabularx}
\usepackage{array}
\usepackage{caption}
\usepackage{svg}
\usepackage{tcolorbox}
\usepackage{rotating} 

\usepackage{enumitem}
\usepackage{comment}
\usepackage{amsfonts} 
\usepackage{makecell}   
\usepackage{xcolor}     
\usepackage{geometry}   

\graphicspath{ {./images/} }

\newcolumntype{L}{>{\raggedright\arraybackslash}X}

\journal{Medical Image Analysis; accepted for publication (DOI: 10.1016/j.media.2026.104083)}

\begin{document}

\begin{frontmatter}



\title{Decoding the Surgical Scene: A Scoping Review of Scene Graphs in Surgery}

\author[tum_cit]{Angelo Henriques\corref{cor1}}
\ead{angelo.henriques@tum.de}

\author[tum_cit]{Korab Hoxha}
\ead{korab.hoxha@tum.de}

\author[tum_hosp]{Daniel Zapp}
\ead{daniel.zapp@mri.tum.de}

\author[tum_hosp]{Peter C. Issa}
\ead{peter.charbelissa@mri.tum.de}

\author[tum_cami]{Nassir Navab}
\ead{nassir.navab@tum.de}

\author[tum_hosp,uoa]{M. Ali Nasseri}
\ead{ali.nasseri@tum.de}

\cortext[cor1]{Corresponding author}

\affiliation[tum_cit]{organization={School of Computation, Information and Technology, Technical University of Munich},
            city={Munich},
            country={Germany}}

\affiliation[tum_hosp]{organization={Klinik und Poliklinik für Augenheilkunde, TUM University Hospital},
            city={Munich},
            country={Germany}}
            
\affiliation[tum_cami]{organization={Computer Aided Medical Procedures, Technical University of Munich},
            city={Munich},
            country={Germany}}
            
\affiliation[uoa]{organization={Department of Biomedical Engineering, University of Alberta},
            city={Edmonton},
            country={Canada}}

\begin{abstract}
As surgical AI transitions from pixel-level detection to complex reasoning, Scene Graphs (SGs) offer the structured, relational representations necessary to decode dynamic surgical environments. This PRISMA-ScR-guided scoping review systematically maps the evolving landscape of SG research in surgery, analyzing 52 primary studies to chart applications and methodological shifts. Our analysis reveals rapid growth, yet uncovers a critical 'data divide': internal-view research (e.g., triplet recognition from endoscopic video) accounts for 81\% of studies and almost exclusively uses real-world 2D video, while external-view operating room modeling relies heavily on simulated data. Methodologically, we identify a decisive shift from foundational graph neural networks to specialized foundation models and generative AI, which together now account for approximately 50\% of research in 2025. Crucially, our synthesis suggests that Scene Graphs are evolving from simple descriptors into essential 'neuro-symbolic guardrails', providing the structured, verifiable intermediate representation needed to prevent hallucinations in increasingly autonomous Surgical Foundation Models. Despite this promise, a major translational gap remains: \textbf{none} of the reviewed studies have proceeded to prospective clinical validation. We conclude that bridging this gap requires moving beyond standard computer vision metrics; we therefore propose the 'Validation Trinity'---prioritizing Semantic Query Success, Latency-Aware Accuracy, and Safety-Critical Recall---as the necessary evaluation framework to bring graph-based surgical AI into clinical practice.
\end{abstract}

\begin{graphicalabstract}
    \centering
    \includegraphics[width=1\textwidth]{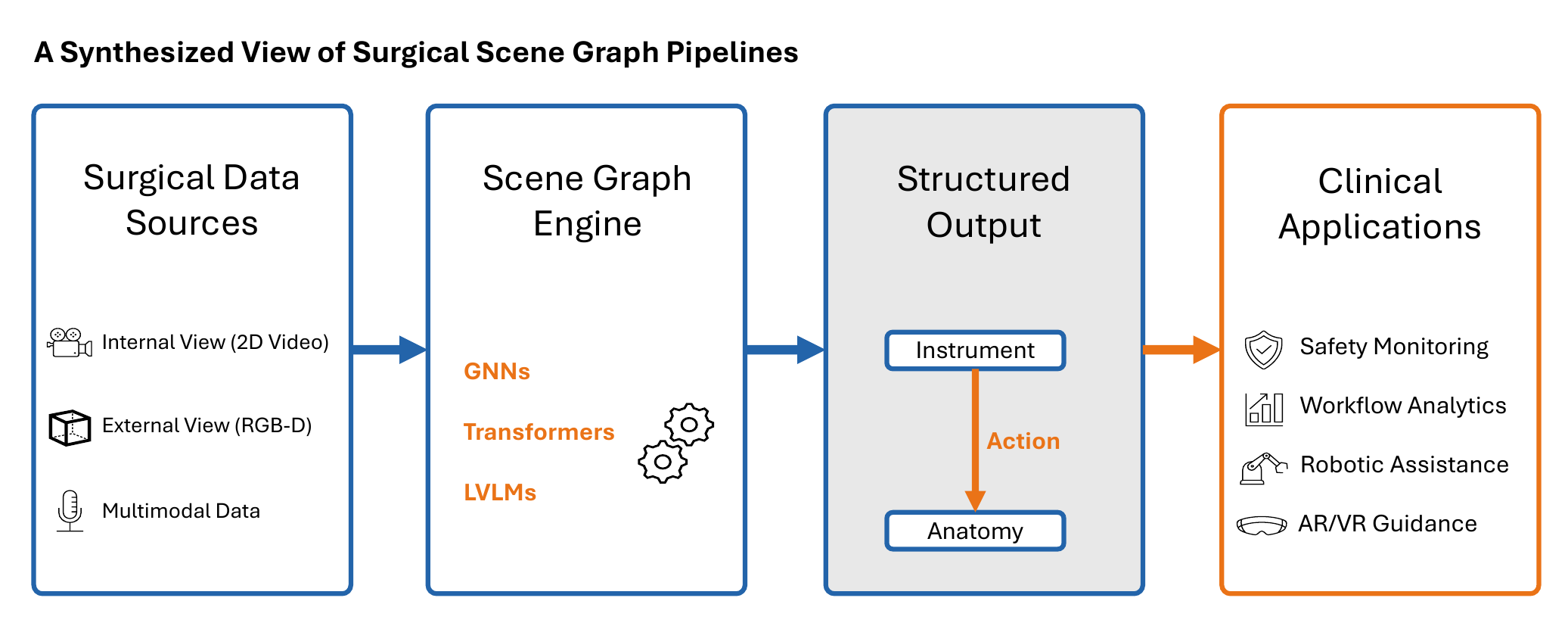}
\end{graphicalabstract}

\begin{highlights}
\item First systematic review to map the rapidly evolving landscape of Surgical Scene Graphs.
\item Quantitative analysis reveals a decisive shift from GNNs to Foundation Models (50\% of 2025 research).
\item Exposes a critical ‘data divide’: internal views use real video, while 4D OR modeling relies on simulation.
\item Scene graphs are emerging as essential neuro-symbolic guardrails to prevent AI hallucinations.
\item Identifying the transition from descriptive analysis to controllable generative surgical simulation.
\end{highlights}


\end{frontmatter}


\section{Introduction}
\label{sec:introduction}

\subsection{Motivation: Why Scene Graphs in Surgery?}
\label{sec:intro_motivation}

Failures in surgical situation awareness, stemming from an inability to accurately interpret the intricate web of interactions within the operating field, are recognized contributors to preventable adverse events \citep{Schulz.2013}. The modern operating room (OR) is a complex, dynamic environment, motivating the pursuit of what \citet{MaierHein.2017} termed ''superhuman'' surgery, which aims to move beyond the data associations that individuals are able to perceive, ''[\dots] into the realm of vast data types and sizes that can only be exploited through modern computing solutions''. Surgeons navigate scenes where multiple instruments maneuver simultaneously and critical anatomical structures lie millimeters from the point of action. While recent advances in surgical data science have made strides in analyzing specific components of this complexity---for instance, identifying instruments or classifying surgical phases---these methods often process elements in isolation. Consequently, the rich relational context that dictates the semantic meaning of the surgical scene remains largely underutilized.

This lack of structured grounding becomes critical in the era of Generative AI. While emerging Large Vision-Language Models (LVLMs) promise 'superhuman' reasoning, these probabilistic models suffer from hallucinations, generating plausible but factually incorrect surgical advice. Here lies the critical utility of the Scene Graph: unlike the opaque latent space of an LLM, a Scene Graph is \textbf{explicit and falsifiable}. It transforms the 'black box' of AI prediction into a 'white box' of verifiable relationships (e.g., confirming a \texttt{<grasper, holding, gallbladder>} edge exists before an autonomous cut is permitted). We ultimately argue that this structured falsifiability positions SGs not merely as a descriptive tool, but as the essential verification layer for future Surgical Foundation Models.

A fundamental challenge lies in capturing and reasoning about the relationships between entities. Understanding a procedure necessitates knowing not just which tools and organs are present, but precisely how they relate: which instrument is dissecting which tissue, or how close a retractor is to a critical nerve \citep{Bigdelou.2011}. While early computational approaches used proxy signals for context, such as instrument co-occurrence \citep{Ahmadi.2006, Twinanda.2016}, these heuristics are often brittle and lack the explicit, compositional structure needed for robust, high-level reasoning. What has been missing is a comprehensive, structured representation of the surgical scene's semantics.

Scene graphs (SGs) offer a powerful paradigm to address this gap. Originating from computer graphics \citep{Strauss.1992} and later formalized for computer vision \citep{Johnson.2015, Tripathi.2019}, a SG is a data structure that encodes entities (e.g., objects, actors) as nodes and their pairwise relationships as directed edges. These relationships are often expressed as triplets in the form of \texttt{<subject, predicate, object>}, representing a visual scene as, for instance, \texttt{<boy, riding, bicycle>}. This paradigm translates powerfully to the surgical domain, where the general triplet becomes clinically critical, e.g. \texttt{<grasper, retracting, gallbladder>}.

Unlike flat lists of detected objects, a SG can explicitly represent and allow querying of complex configurations, enabling transformative applications. These include automated intra-operative safety checks (e.g., monitoring tool-tissue proximity), objective skill assessment via interaction dynamics, context-aware robotic and augmented reality assistance, and the generation of detailed, queryable logs for reporting and training. The recent convergence of sophisticated deep learning models for perception, the increasing availability of large-scale surgical video datasets, and the maturation of relational reasoning architectures makes the automatic construction and utilization of surgical SGs technically feasible now more than ever before.

\subsection{Historical Context and Scope of this Review}
\label{sec:intro_history}

The concept of graphically representing scenes has a rich history. Scene graphs were first formalized in computer graphics to manage and render complex 3D environments efficiently \citep{Strauss.1992, Hughes.2013}. Their modern application in AI was catalyzed in the mid-2010s with their transition into computer vision, where the focus shifted to semantic understanding \citep{Johnson.2015}. The release of large-scale annotated datasets like Visual Genome \citep{Krishna.2017} spurred research into automatically generating SGs from images, proving their value for downstream tasks like captioning and visual question answering \citep{Tripathi.2019, Dhamo.2020}.

\begin{figure}[htbp]
    \centering
    \includegraphics[width=1\textwidth]{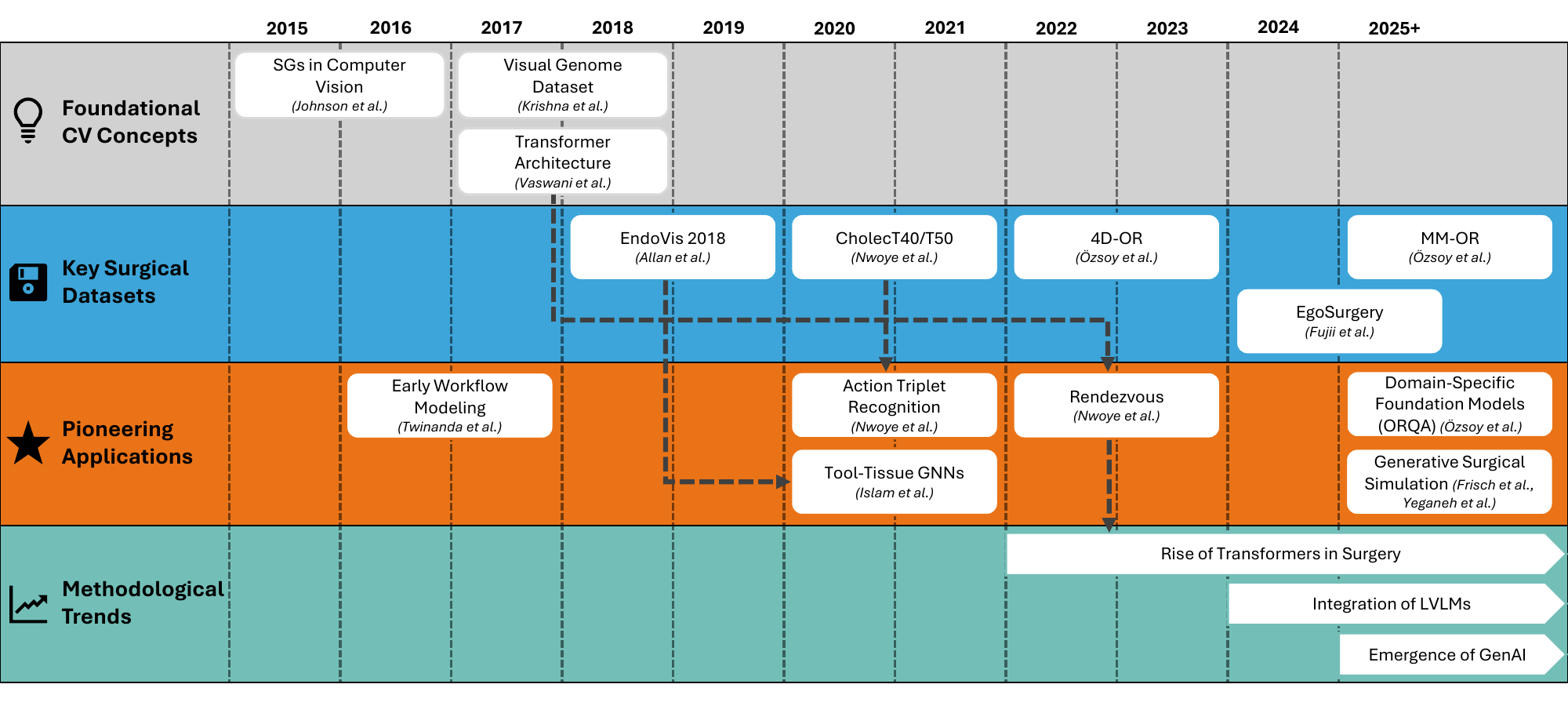}
    \caption{\textbf{The Evolving Landscape of Surgical Scene Graphs.} This multi-layered timeline illustrates the parallel advancements that have shaped the field from 2015 to 2025. \textbf{(Top Lane)} Foundational concepts from general computer vision provided the core technologies. \textbf{(Middle-Top Lane)} The release of key surgical datasets enabled new research directions. \textbf{(Middle-Bottom Lane)} These culminated in pioneering surgical applications and methodological breakthroughs. \textbf{(Bottom Lane)} These advances reflect and have been driven by overarching methodological trends in the field. Dashed arrows indicate direct conceptual or data-driven influence between key milestones.}
    \label{fig:timeline}
\end{figure}

Within surgery, while early work laid the groundwork for workflow analysis \citep{Ahmadi.2006, Twinanda.2016}, the explicit application of graph-based relational modeling is a recent development. As illustrated in the multi-layered timeline in Figure~\ref{fig:timeline}, dedicated research in this area emerged around 2020, building upon \textbf{foundational concepts} from computer vision like the Transformer architecture \citep{Vaswani.2017}. This progress was catalyzed by the release of \textbf{key surgical datasets}. For example, new annotations on the EndoVis 2018 challenge dataset \citep{Allan.2020} directly enabled the first \textbf{pioneering applications} of GNNs to model tool-tissue interactions \citep{Islam.2020}. Similarly, the release of the CholecT40/T50 benchmarks \citep{Nwoye.2020, Nwoye.2022} spurred the formalization of the action triplet recognition task. This methodological evolution is exemplified by the influential Rendezvous architecture \citep{Nwoye.2022}, which vividly illustrates the 'Rise of Transformers in Surgery' \textbf{methodological trend} by applying the Transformer paradigm to this task.

This evolution, from early workflow modeling to the most recent integration of domain-specific foundation models \citep{Ozsoy.2025b} and the novel application of SGs in generative AI for controllable surgical simulation \citep{Frisch.2025, Yeganeh.2025}, reflects a field rapidly advancing towards more holistic and intelligent systems. This historical trajectory provides the essential scientific context for this review's exploration of how scene graphs are currently being applied and are poised to shape the future of surgical data science.

\subsection{Gaps in Existing Reviews and Novelty of this Work}
\label{sec:intro_gaps}

Despite growing interest, a consolidated overview focused on the surgical domain is currently lacking. Existing reviews are either broad technical surveys on graph-based methods that omit domain-specific surgical complexities \citep{Chang.2023, Chen.2024, Li.2024}, or are surgical AI reviews that focus on specific downstream applications like action recognition \citep{Bian.2024}, workflow recognition \citep{Demir.2023}, and visual question answering \citep{Ding.2025}, rather than the graph representation itself. Other key reviews predate the recent uptake of explicit graph-based modeling \citep{Garrow.2021, Moglia.2021}. This creates a critical gap for researchers and clinicians seeking a unified resource on this interdisciplinary topic.

The novelty of this scoping review lies in presenting the first systematic mapping of the literature exclusively dedicated to surgical scene graph applications. Given the field's rapid growth and convergence with transformative AI, such a synthesis is both timely and necessary. Specifically, we:
\begin{itemize}
\item Provide a comprehensive, structured reference charting the state-of-the-art.
\item Identify key patterns and inconsistencies in current methodologies and application trends.
\item Highlight underexplored areas and specific opportunities for innovation.
\item Establish a clear baseline to guide and accelerate future research towards clinical translation.
\end{itemize}
In essence, this review aims to narrate the evolving story of surgical scene graphs, clarifying their current role and potential trajectory in transforming surgical AI.

\subsection{Objectives and Research Questions}
\label{sec:intro_objectives}

The primary objective of this scoping review is to systematically map the current landscape of research applying scene graph methodologies within surgical contexts. We aim to synthesize existing knowledge, identify key advancements and persistent challenges, and outline actionable directions for future work. To achieve this, we address the following research questions:
\begin{enumerate}
\item \textbf{Applications:} What specific surgical domains and environments have surgical scene graphs been applied to? What problems are researchers trying to solve using this representation?
\item \textbf{Methodologies:} How are surgical scene graphs typically defined? What data sources, modalities, and computational methods are employed for graph construction and reasoning?
\item \textbf{Benefits and Challenges:} What are the reported benefits of using scene graphs? What are the main limitations and challenges encountered in their construction and application?
\item \textbf{Future Directions:} Based on current trends and identified gaps, what are the primary emerging trends and predicted trajectories for future research?
\end{enumerate}
By addressing these questions, this review seeks to illuminate the motivations, implementations, and outcomes of SG adoption in surgical data science and delineate a clear trajectory for future work.

\section{Background}
\label{sec:background}

This section defines the key characteristics that distinguish surgical scene graphs from their general-purpose counterparts and introduces a formal taxonomy used to categorize the works analyzed in this review.

\subsection{Distinctiveness of Surgical Scene Graphs}
\label{sec:distinctiveness}

While surgical scene graphs share the same underlying graph formalism as those in general computer vision, the clinical domain imposes specialized semantics, stricter performance demands, and unique constraints.

\paragraph{Domain-Specific Semantics and Knowledge Integration} A primary distinction of surgical scene graphs lies in their highly specialized semantics. Nodes represent specific instruments (e.g., monopolar hook, ProGrasp forceps) and fine-grained anatomical structures, while edges denote precise clinical actions (e.g., "dissects", "retracts", "sutures") that carry significant clinical meaning. These relationships are governed by procedural logic, anatomical constraints, and codified safety rules, such as the Critical View of Safety. Consequently, effective surgical scene graph models often benefit from or require the integration of prior knowledge, which may be derived from medical ontologies, surgical atlases, or procedural checklists \citep{Cao.2021, Ban.2024, Dsouza.2023}. Such knowledge fusion is often essential for accurate interpretation and reasoning beyond purely data-driven pattern recognition.

\paragraph{Unique Data and Performance Challenges} The surgical environment poses a unique set of technical and data-related hurdles. These include severe class imbalances in datasets, stringent requirements for accuracy and reliability due to patient safety implications, the need for real-time performance in many applications, and significant visual noise from factors like smoke and blood. Furthermore, model generalization is challenged by high variability across patients, surgeons, and institutions. A detailed discussion of these challenges and their impact on the field is provided in Section \ref{sec:limitations}.

\subsection{Taxonomy of Scene Graphs in Surgery}
\label{sec:taxonomy}

\begin{figure}[htbp]
    \centering
    \includegraphics[width=1\textwidth]{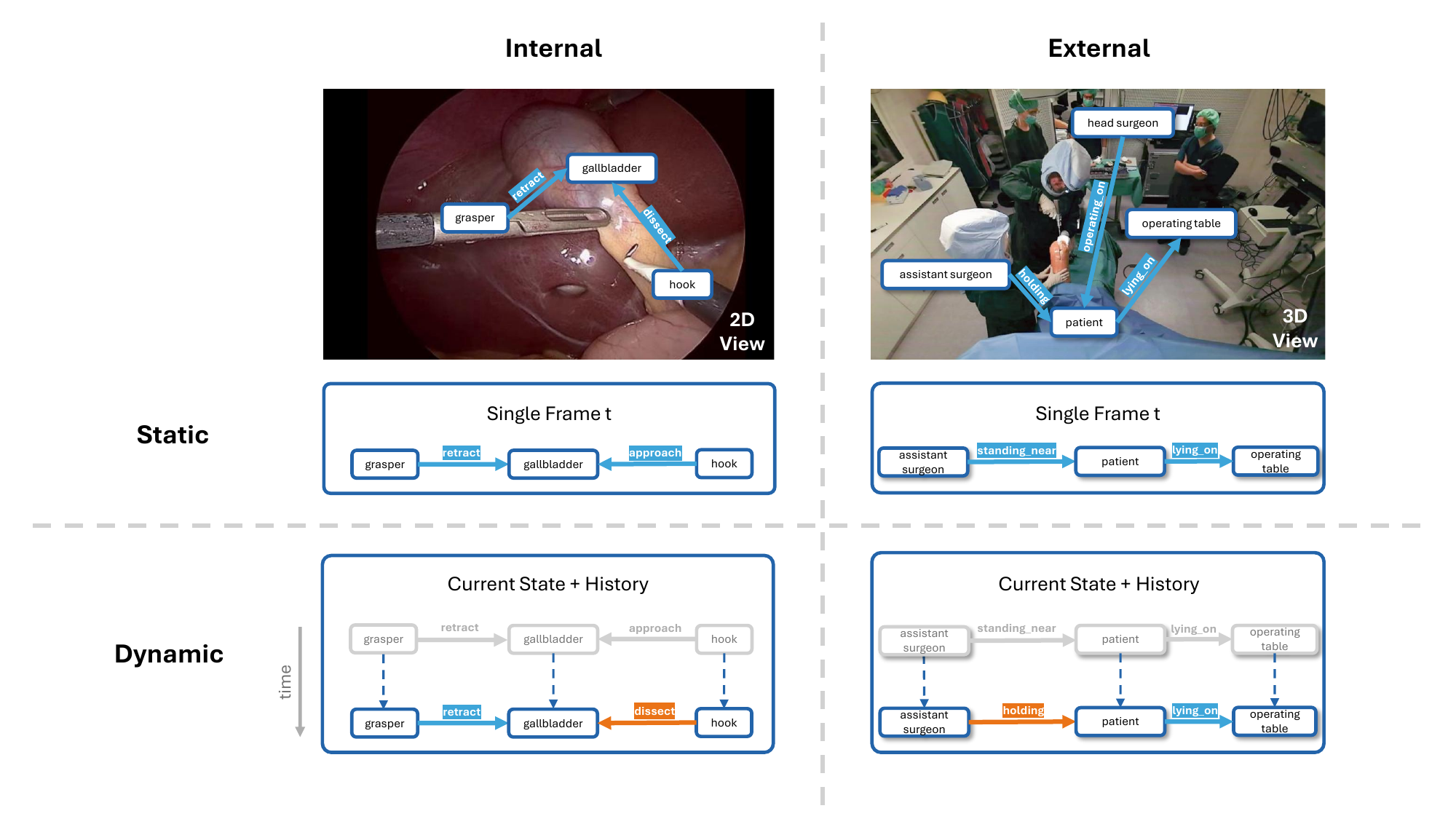}
    \caption{\textbf{Visualization of the Surgical Scene Graph Taxonomy across Viewpoint, Temporality, and Dimensionality.} The columns distinguish between the \textbf{Internal (2D)} and \textbf{External (3D)} viewpoints. 
    \textbf{(Top Row)} The \textit{Static Scene Graph} depicts a snapshot at a single frame. To distinguish dimensionality, Internal nodes (left) are rendered flat (2D), while External nodes (right) feature drop shadows to imply volumetric 3D positioning.
    \textbf{(Bottom Row)} The \textit{Dynamic Scene Graph} illustrates temporal evolution. The faded nodes represent the history at time $t$, connected by dashed vertical lines (temporal tracking) to the solid nodes at time $t+n$ (current state). The \textbf{Orange} edge highlights the semantic state transition (e.g., from \texttt{approach} to \texttt{dissect}) captured by the dynamic graph.}
\label{fig:taxonomy_panel}
\end{figure}

To consistently compare existing studies, we group surgical scene graphs along three primary axes---time, dimensionality, and camera viewpoint---which are summarized below and illustrated in Figure \ref{fig:taxonomy_panel}.

\paragraph{Temporal Aspect: Static vs. Dynamic}
\begin{itemize}
\item \textbf{Static:} Represents a single snapshot in time, typically from one image frame. These are suitable for analyzing instantaneous spatial configurations or co-occurring entities \citep{Islam.2020}.
\item \textbf{Dynamic (Spatio-Temporal):} Represents relationships and entity states as they evolve over time. These are essential for modeling surgical actions, procedural workflows, and interaction histories \citep{Holm.2023, Ozsoy.2023}.
\end{itemize}

\paragraph{Dimensionality: 2D vs. 3D/4D}
\begin{itemize}
\item \textbf{2D:} Constructed from standard planar images (e.g., endoscopic video). Relationships are defined within the 2D image plane \citep{Islam.2020, Holm.2023}.
\item \textbf{3D/4D:} Entities are localized in 3D space, typically requiring depth sensors (e.g., RGB-D). This enables modeling of real-world spatial relationships like metric distances \citep{Ozsoy.2022}. When tracked over time, this becomes a \textbf{4D} (3D + time) representation, capturing the complete spatio-temporal evolution of the environment \citep{Ozsoy.2023}.
\end{itemize}

\paragraph{Viewpoint: Internal vs. External}
\begin{itemize}
\item \textbf{Internal:} Focuses on the operative field, captured via an endoscope, laparoscope, or microscope. Nodes include instruments and anatomy, while edges represent surgical actions (e.g., cutting, grasping) and tool-tissue interactions \citep{Islam.2020, Nwoye.2020, Holm.2023}. These are used for action recognition, safety assessment, and workflow analysis.
\item \textbf{External:} Models the broader operating room, captured from cameras positioned outside the patient. Nodes include staff, equipment, and the patient, while edges represent team interactions, equipment usage, and spatial arrangements \citep{Ozsoy.2022, Ozsoy.2023, Ozsoy.2024b}. These are used for analyzing OR workflow, team coordination, and room state.
\end{itemize}

It is important to note that these categories are not mutually exclusive; for instance, a study might utilize a dynamic 4D external scene graph \citep{Ozsoy.2023}. Understanding these distinctions is key to interpreting the capabilities and limitations of different surgical scene graph methodologies reported in the literature.

\subsection{Defining the Scope: From Triplets to Scene Graphs}
\label{sec:scope_definition}
A critical distinction in this review is the relationship between Surgical Action Triplets and full Surgical Scene Graphs. We define a \textbf{Surgical Action Triplet} as the atomic unit of a scene graph, consisting of a single directed edge \texttt{<Instrument, Action, Target>} (e.g., \texttt{<Grasper, Retracting, Gallbladder>}). While some studies focus exclusively on recognizing these isolated triplets without constructing a connected graph of the entire scene, we include them in this scoping review for two reasons:
\begin{enumerate}
    \item \textbf{Foundational Relevance:} Triplet recognition represents the core mechanism of relational modeling. The computational challenges involved (e.g., relating spatially disjoint entities, handling interaction dynamics) are identical to those required for full scene graph generation.
    \item \textbf{Evolutionary Trajectory:} As evidenced in our results, the field has historically evolved from recognizing single interactions (triplets) in 2D endoscopic video to generating dense, multi-relational graphs in 4D environments.
\end{enumerate}
Consequently, we exclude studies that perform purely pixel-level segmentation or object detection unless they explicitly model the \textit{relationships} between entities, either as triplets or connected graphs.

\section{Methodology}
\label{sec:methodology}

This scoping review was conducted and reported in accordance with the \emph{Preferred Reporting Items for Systematic Reviews and Meta-Analyses extension for Scoping Reviews} (PRISMA-ScR) checklist \citep{Tricco.2018}. The methodology was designed to systematically map the breadth and depth of research applying scene graph concepts to surgical contexts, identify key characteristics, and summarize the evolution, trends, and gaps in this rapidly emerging field. The study protocol was prospectively registered on the \emph{Open Science Framework} (OSF; \url{osf.io/jh68z}). The final title of the manuscript was refined from the original registered title to better reflect the synthesized findings and overall scope of the work.

\subsection{Scoping Review Framework Justification}
\label{sec:methodology_justification}
\begin{figure}[htbp]
    \centering
    \includegraphics[width=1\textwidth]{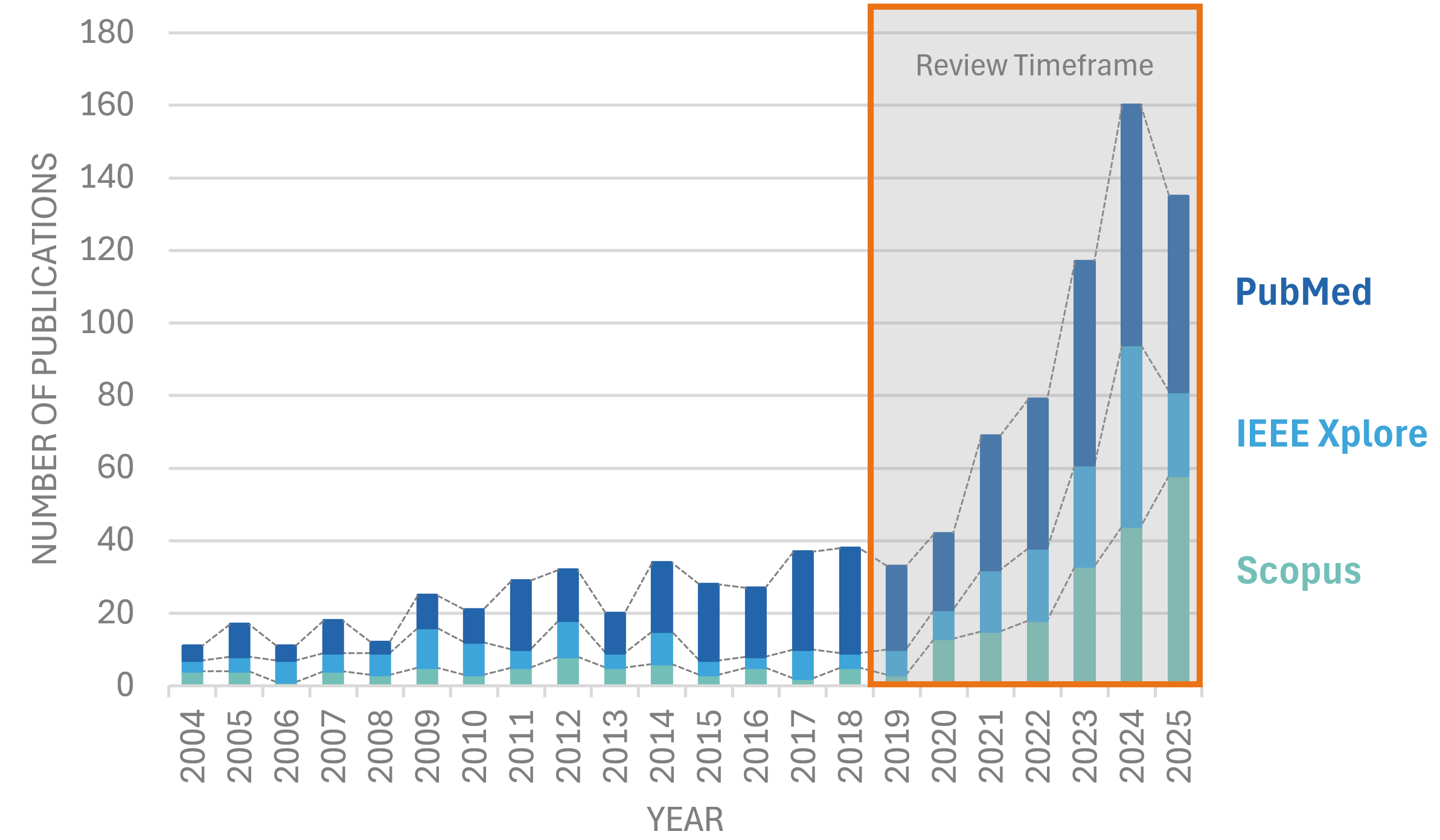}
    \caption{\textbf{Annual Publication Count matching the Core Search String.} This illustrates the recent growth of interest in work on surgical scene graphs. Data sources: Scopus, IEEE Xplore, PubMed. The bar for 2025 includes studies published up to the final search date of July 29th, 2025.} 
    \label{fig:pub_timeline}
\end{figure}

A scoping review is well-suited to nascent, heterogeneous fields where the literature spans multiple venues and methodologies. Scene graphs entered surgical data science only recently, publications rise steeply after 2019 (Figure~\ref{fig:pub_timeline}), and cover diverse applications. Unlike a systematic review that targets a narrowly defined outcome, the scoping approach enables a comprehensive, descriptive map of how the field has evolved.

\subsection{Eligibility Criteria}
\label{sec:methodology_criteria}

Studies were \emph{included} if they met the following criteria:
\begin{itemize}[leftmargin=*]
    \item \textbf{Publication Status}: Peer-reviewed articles were prioritized. However, given the rapid pace of the field, seminal preprints (e.g., recent foundation models from 2025--2026) were included if they represented the state-of-the-art. These exceptions are explicitly marked in the supplementary online materials.
    \item \textbf{Time Frame:} Published between January 1st, 2019, and July 29th, 2025 (date of final search).
    \item \textbf{Content:} Explicitly developed or utilized techniques based on scene graphs or the closely related concept of surgical action triplet recognition (\texttt{<instrument, action, target>}).
    \item \textbf{Context:} Applied these techniques within a surgical context (e.g., operative video, simulated surgery, operating-room footage).
    \item \textbf{Language:} Published in English.
\end{itemize}
Studies were \emph{excluded} if they were: reviews, editorials, abstracts without sufficient detail, non-peer-reviewed preprints (unless specifically identified as seminal/highly relevant by experts), studies where scene graphs or triplets were mentioned only incidentally without forming a core part of the methodology or investigation, or non-surgical medical imaging studies without direct relevance to surgical procedures or environments.

\subsection{Information Sources and Search Strategy}
\label{sec:methodology_search}

A systematic search was performed across three electronic databases: PubMed, Scopus, and IEEE Xplore. To ensure the review's timeliness in this rapidly evolving field, the search was conducted in two phases. An initial search was performed in September 2024, followed by a formal update in July 2025 to capture the most recent publications. 

The identical search string was used in both search phases. It combined terms related to scene graphs/triplets with terms related to surgery:
\vspace{0.3em}
\begin{verbatim}
((scene AND graph) OR "triplet recognition")
AND
(surg* OR medical OR "operation room" OR "OR")
\end{verbatim}
\vspace{0.3em}

This primary database search was supplemented by:
\begin{itemize}[leftmargin=*]
    \item \textbf{Keyword Search in Google Scholar:} The core search terms were also queried directly within Google Scholar. Due to the high volume of results generated by this platform, Google Scholar's built-in filters were first applied to restrict results to the specified timeframe (2019--2024). Subsequently, the first 500 unique results from this filtered list were manually screened based on title and abstract for potential inclusion, acknowledging that relevance typically diminishes rapidly beyond the initial pages of results.
    \item \textbf{Citation Searching:} Backward (screening reference lists of included studies) and forward (tracking citations of included studies) citation searching using Google Scholar.
    \item \textbf{Manual Identification:} Screening the reference lists of relevant existing reviews identified during the search process.
    \item \textbf{Expert Consultation:} Incorporating recommendations from leading researchers in the field, identified through prominent publications and conference involvement, to capture potentially missed key or recent works. All recommendations were screened against the full eligibility criteria. This supplementary search also allowed for the inclusion of highly relevant works (including preprints like \citep{Ozsoy.2025c}). 
\end{itemize}

\subsection{Selection of Sources of Evidence (Screening)}
\label{sec:methodology_screening}

\begin{figure}[htbp]
    \centering
    \includegraphics[width=0.7\textwidth]{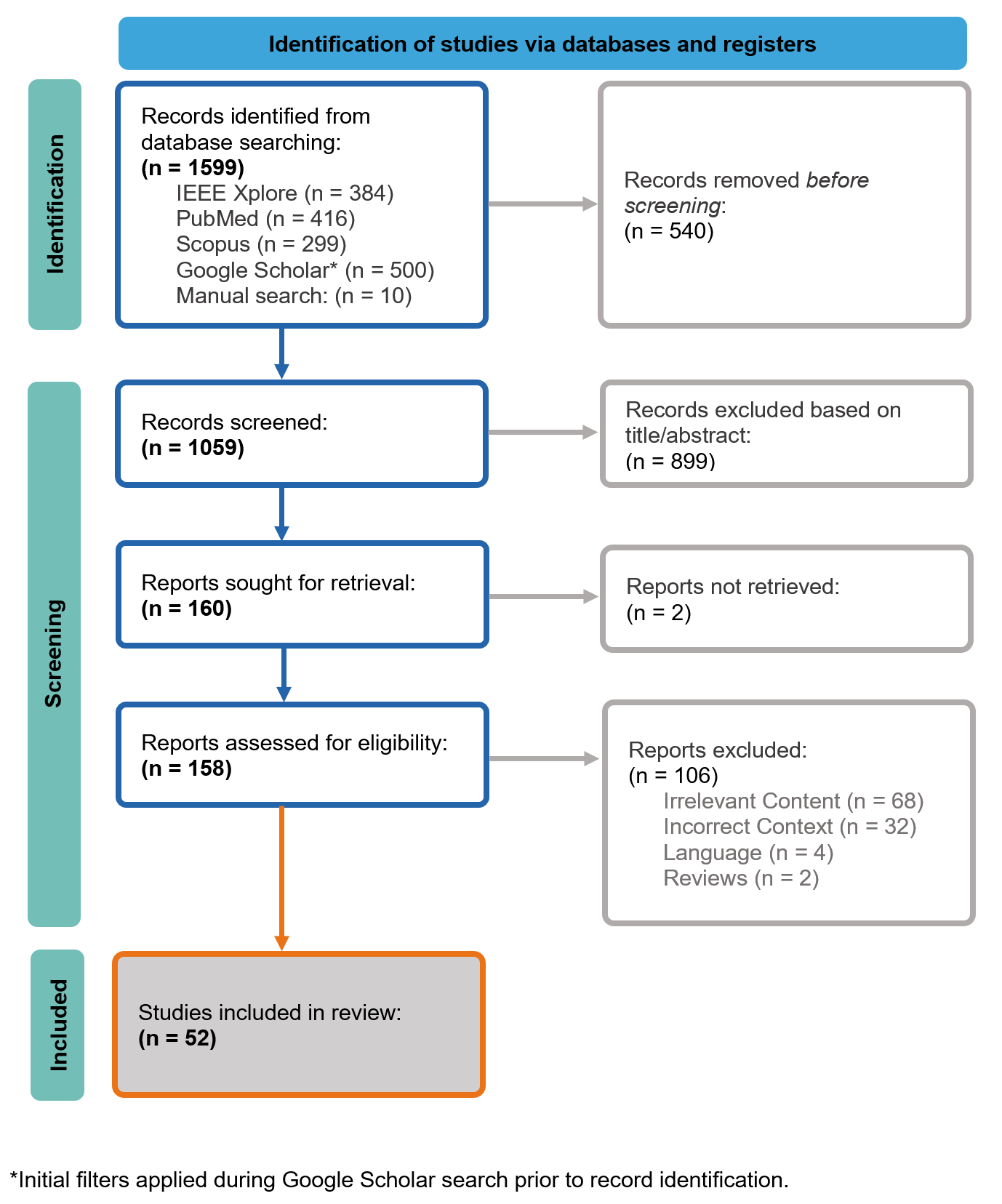} 
    \caption{\textbf{PRISMA Flow Diagram} \citep{Tricco.2018}. It illustrates the identification, screening, and inclusion process of the combined original and updated systematic searches.}
    \label{fig:prisma}
\end{figure}

The selection process involved two sequential stages:
\begin{enumerate}[leftmargin=*]
    \item \textbf{Title and Abstract Screening:} Citations retrieved from the searches were imported into Citavi.
    After deduplication, two reviewers (A.H., K.H.)
    independently screened the titles and abstracts against the eligibility criteria. Any disagreements regarding inclusion or exclusion were resolved through discussion between the two reviewers or, if consensus could not be reached, through consultation with a third reviewer (A.N.). 
    \item \textbf{Full-Text Review:} The same reviewers independently assessed each potentially eligible article in full. Disagreements were resolved using the procedure above.
\end{enumerate}
The detailed flow of study identification, screening, eligibility assessment, and final inclusion is documented in the PRISMA 2020 flow diagram (Figure~\ref{fig:prisma}).

\subsection{Data Charting (Extraction)}
\label{sec:methodology_extraction}

A data charting form was designed to capture key information relevant to the scoping review's objectives and research questions. The form was developed iteratively and piloted on a small subset of included studies. Data extraction was performed by one reviewer (A.H.) and subsequently verified by a second reviewer (K.H.) to ensure accuracy. In cases where a study addressed multiple tasks (e.g., both triplet recognition and workflow analysis), the primary application was determined based on the task receiving the most extensive experimental validation. The extracted data items included: Author(s), Year, Publication Title, Surgical Domain, Scene Graph Type (using the taxonomy from Section~\ref{sec:taxonomy}), Core Construction Method(s), Data Modalities Used, Dataset(s) Used, Key Contributions, Quantitative Lift, and reported Limitations. 

\subsection{Synthesis of Results}
\label{sec:methodology_synthesis}

The extracted data were synthesized narratively to provide a comprehensive overview of the field. Findings were grouped thematically to address the specific research questions outlined in Section~\ref{sec:intro_objectives}. Specifically, synthesis focused on summarizing the range of application domains and tasks (addressing RQ1), identifying common and emerging methodological approaches for graph definition, construction, and reasoning (addressing RQ2), and compiling the reported benefits alongside persistent challenges and limitations (addressing RQ3). A chronological perspective was adopted where appropriate to illustrate the field's evolution, mapping the trajectory from early work towards current trends (e.g., integration with LLMs/LVLMs, multimodal inputs, exploration of egocentric views) to inform the discussion on future directions (addressing RQ4). Key findings are presented textually and complemented by detailed summary tables in \ref{sec:appendix_tables} (particularly Table~\ref{tab:summary_internal_related_split} and Table~\ref{tab:summary_external_split}).

\section{Scene Graph Construction Methods}
\label{sec:construction_methods}

Constructing a surgical scene graph from raw sensor data is a multi-stage process, as conceptualized in Figure \ref{fig:pipeline}. The process begins with raw input data, which is processed by a \textbf{Perception Module} to identify scene entities (nodes) and extract their features. A subsequent \textbf{Graph Inference Module} then reasons about the relationships between these entities to predict the final graph structure (edges). This structured output then serves as a semantic foundation for various downstream applications. This section details the datasets, computational methods, data sources and modalities that underpin this pipeline.

\subsection{Data Sources and Modalities}
\label{sec:data_sources}

The choice of data source directly dictates the dimensionality and complexity of the resulting scene graph. The most common modality, particularly for internal-view analysis, is standard \textbf{2D RGB video} from endoscopes, laparoscopes, or surgical microscopes \citep{Islam.2020, Nwoye.2022, Yuan.2024}. Processing these video streams sequentially allows models to leverage temporal context to build dynamic scene graphs and understand actions over time \citep{Holm.2023}.

For applications requiring true 3D spatial understanding, such as external OR modeling or robotics, \textbf{RGB-D data} is increasingly employed. The depth channel is vital for extracting metric 3D positions and real-world spatial relationships, enabling the construction of 4D (3D + time) scene graphs that track dynamic spatial interactions \citep{Ozsoy.2023}. The latest trend is toward integrating diverse streams into \textbf{multimodal} frameworks. Datasets like MM-OR, for instance, fuse multi-view RGB-D video with audio, speech transcripts, and robotic system logs to enable more robust and comprehensive scene understanding by leveraging complementary information \citep{Ozsoy.2025}. Finally, some approaches use auxiliary data streams, like pre-defined phase annotations or instrument logs, as weak supervision signals to guide graph generation \citep{Koksal.2024}.

\subsection{Available Surgical Datasets}
\label{sec:available_datasets}

Progress in the field is intrinsically linked to the availability of specialized datasets, which have evolved from repurposed videos to large, purpose-built multimodal benchmarks. A comparative summary of key resources is provided in Tables \ref{tab:datasets_internal} and \ref{tab:datasets_external}. While scene graph datasets exist for diagnostic radiology, such as for report generation from chest X-rays \citep{Johnson.2019, Wu.2021} or hemorrhage detection in CT \citep{Sanner.2024}, these are excluded from this review to maintain a strict focus on the surgical environment.

The first wave of research augmented existing datasets, such as the EndoVis 2018 challenge data, with new tool-tissue interaction annotations, which was foundational for early internal Scene Graph Generation (SGG) \citep{Islam.2020, Allan.2020}. A significant catalyst for fine-grained interaction analysis was the release of \textbf{CholecT40}, \textbf{CholecT45}, and the benchmark \textbf{CholecT50} \citep{Nwoye.2020, Nwoye.2022}, which provided the first dense action triplet annotations and spurred rapid methodological development through associated challenges \citep{Nwoye.2023, Nwoye.2023b}. Other key internal-view resources include CATARACTS, used for graph-based workflow analysis \citep{Holm.2023, Koksal.2024}, and custom datasets for studying domain adaptation and safety assessment \citep{Seenivasan.2022, Murali.2024}.

To model the broader surgical context, the \textbf{4D-OR} dataset was introduced as the first major public benchmark for external OR analysis, providing 4D (multi-view RGB-D + time) data that enabled research into holistic 3D/4D reasoning \citep{Ozsoy.2022}. The most recent datasets push towards richer inputs, with \textbf{MM-OR} representing the state-of-the-art for multimodal OR understanding \citep{Ozsoy.2025}, and \textbf{SSG-VQA} linking scene graphs to language by providing a benchmark for surgical visual question answering \citep{Yuan.2024}. Overall, the trajectory shows a clear progression from 2D real-world data towards simulated 3D/4D and multimodal data, highlighting a persistent need for more large-scale, richly annotated real-world datasets.

\begin{table*}[htbp]
    \centering
    \caption{\textbf{Key Datasets for Internal View Surgical Scene Graph Research.} Superscripts in the first column refer to the primary citation(s) listed below. *Year refers to the year the dataset became relevant for SG research. `Data' indicates Real-world vs. Simulated sources. \textbf{Vol.} denotes dataset volume (Videos unless specified).}
    \label{tab:datasets_internal}
    \scriptsize 
    \setlength{\tabcolsep}{2.5pt} 
    \renewcommand{\arraystretch}{1.2} 
    
    \begin{tabular}{@{} 
        >{\raggedright\arraybackslash}p{0.20\textwidth} 
        >{\raggedright\arraybackslash}p{0.14\textwidth} 
        >{\raggedright\arraybackslash}p{0.05\textwidth} @{\hspace{1em}} 
        >{\raggedright\arraybackslash}p{0.06\textwidth} 
        >{\raggedright\arraybackslash}p{0.27\textwidth} 
        >{\raggedright\arraybackslash}p{0.21\textwidth} 
    @{}}
        \toprule
        \textbf{Dataset [Year*]} & \textbf{Procedure} & \textbf{Data} & \textbf{Vol.} & \textbf{Key Annotation} & \textbf{SG Application} \\
        \midrule
        
        \textbf{EndoVis 2018} \newline \textbf{+ Annots}$^{1}$ [2020] & Rob. Nephrectomy & Real & 19 & Tool-Tissue Interactions (Graph), Seg. & Early supervised internal SGG. \\
        \addlinespace
        
        \textbf{CholecT40}$^{2}$ [2020] & Lap. Cholec. & Real & 40 & Action Triplets (\texttt{<I,V,T>}) & Foundational dataset for triplet recognition. \\
        \addlinespace
        
        \textbf{CholecT45 / T50}$^{3}$ [2022] & Lap. Cholec. & Real & 45/50 & Action Triplets (\texttt{<I,V,T>}) & Benchmark for action triplet recognition. \\
        \addlinespace
        
        \textbf{Custom TORS / Nephr.}$^{4}$ [2022] & Rob. / Lap. & Real & 14/22 & Triplets, Phase, Instruments & Domain adaptation for interaction recognition. \\
        \addlinespace
        
        \textbf{CATARACTS} \newline \textbf{/ CAT-SG}$^{5,11}$ [2023] & Cataract (Microscope) & Real & 50 & Phase, Instruments (Base); Dynamic SG & SG-based workflow \& phase analysis. \\
        \addlinespace
        
        \textbf{Endoscapes+} \newline \textbf{/ Custom CVS}$^{6}$ [2023] & Lap. Cholec. & Real & 201 & BBox/Seg., CVS Assessment & Latent SGs for surgical safety (CVS). \\
        \addlinespace
        
        \textbf{EndoVis-RS17} \newline \textbf{/ RS18}$^{7}$ [2023] & Rob. Surgery & Real & 31 & Referring Expressions (Text + Mask) & Referring instrument segmentation (VL). \\
        \addlinespace
        
        \textbf{SVL (SurgVLP)}$^{8}$ [2023] & Lap. Surgery (Education) & Real & 157k Clips & ASR Transcripts (for VLP) & Large-scale surgical VLP foundation. \\ 
        \addlinespace
        
        \textbf{SSG-VQA}$^{9}$ [2024] & Lap. Cholec. & Real & 80 & Synth. QA Pairs from Scene Graphs & SG integration in surgical VQA. \\
        \addlinespace
        
        \textbf{Endoscapes-SG201}$^{10}$ [2025] & Lap. Cholec. & Real & 201 & Holistic SG (Hand ID, Action, Target) & Holistic scene understanding. \\
        
        \bottomrule
    \end{tabular}
    \begin{flushleft} 
    \scriptsize 
    \textbf{References:}
    $^{1}$\citep{Allan.2020, Islam.2020},
    $^{2}$\citep{Nwoye.2020},
    $^{3}$\citep{Nwoye.2022},
    $^{4}$\citep{Seenivasan.2022},
    $^{5}$\citep{AlHajj.2019},
    $^{6}$\citep{Murali.2023, Murali.2024},
    $^{7}$\citep{Wang.2024},
    $^{8}$\citep{Yuan.2023},
    $^{9}$\citep{Yuan.2024},
    $^{10}$\citep{Shin.2025},
    $^{11}$\citep{Holm.2025}. 
    \end{flushleft}
\end{table*}

\begin{table*}[htbp]
    \centering
    \caption{\textbf{Key Datasets for External View Surgical Scene Graph Research.} Superscripts in the first column refer to the primary citation(s) listed below. *Year refers to the year the dataset became relevant for SG research. `Data' indicates Real-world vs. Simulated source material. \textbf{Vol.} denotes dataset volume (Videos unless specified).}
    \label{tab:datasets_external}
    \scriptsize 
    \setlength{\tabcolsep}{2.5pt} 
    \renewcommand{\arraystretch}{1.2} 
    
    \begin{tabular}{@{} 
        >{\raggedright\arraybackslash}p{0.20\textwidth} 
        >{\raggedright\arraybackslash}p{0.14\textwidth} 
        >{\raggedright\arraybackslash}p{0.05\textwidth} @{\hspace{1em}} 
        >{\raggedright\arraybackslash}p{0.06\textwidth} 
        >{\raggedright\arraybackslash}p{0.27\textwidth} 
        >{\raggedright\arraybackslash}p{0.21\textwidth} 
    @{}}
        \toprule
        \textbf{Dataset [Year*]} & \textbf{Procedure} & \textbf{Data} & \textbf{Vol.} & \textbf{Key Annotation} & \textbf{SG Application} \\
        \midrule
        
        \textbf{MVOR}$^{1}$ [2021] & Vertebroplasty \& Biopsy & Real & 4 Days & Sparse labels (Proof-of-Concept) & Precursor to 4D-OR / MM-OR. \\
        \addlinespace
        
        \textbf{4D-OR}$^{2}$ [2022] & Knee Arthroplasty & Sim. & 10 & 3D BBox, Poses, 4D Scene Graphs & Foundational benchmark for external 4D SGG. \\
        \addlinespace
        
        \textbf{MM-OR}$^{3}$ [2025] & Rob. Knee Arthroplasty & Sim. & 39 & Dense SGG, Seg., Multimodal labels & State-of-the-art benchmark for multimodal OR. \\
        
        \bottomrule
    \end{tabular}
    \begin{flushleft}
    \scriptsize
    \textbf{References:}
    $^{1}$\citep{Srivastav.2018},
    $^{2}$\citep{Ozsoy.2022},
    $^{3}$\citep{Ozsoy.2025}.
    \end{flushleft}
\end{table*}

\subsection{Computational Methods}
\label{sec:computational_methods}

\begin{figure}[htbp]
    \centering
    \includegraphics[width=1\textwidth]{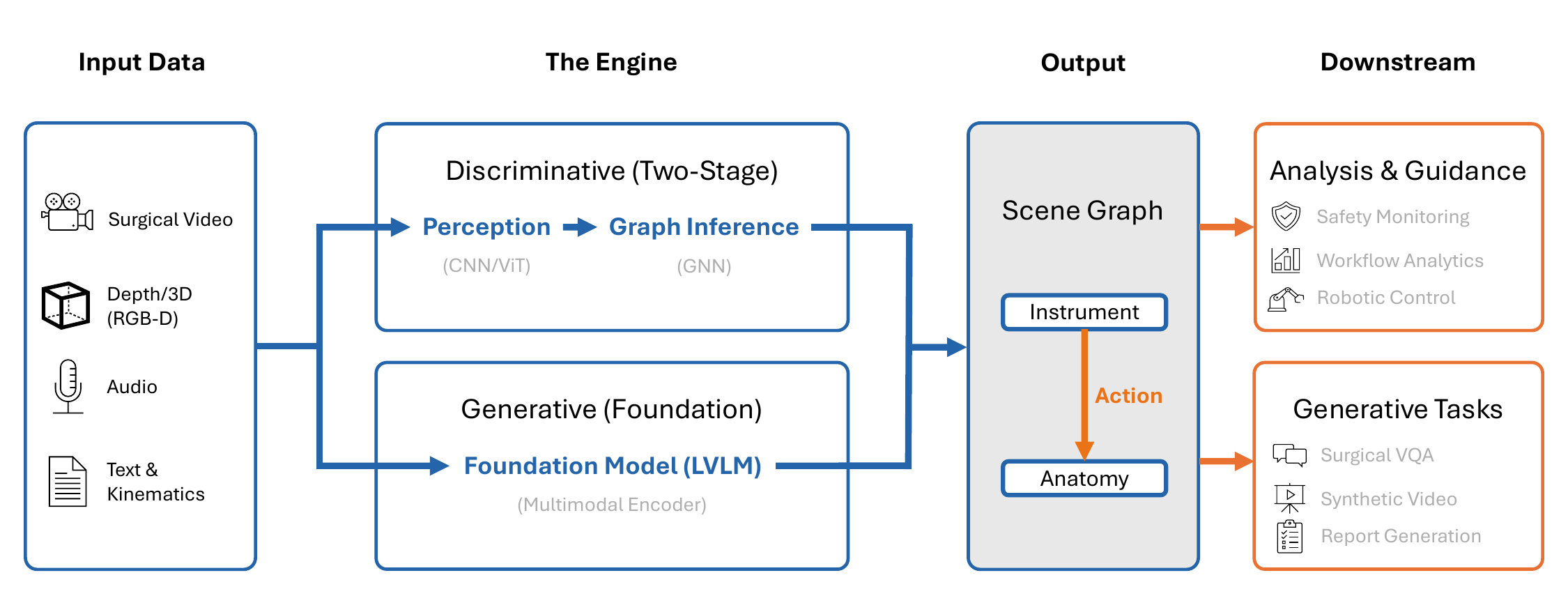} 
    \caption{\textbf{Evolution of Surgical Scene Graph Construction Pipelines.} 
    To address increasing data complexity, methodologies have bifurcated into two paradigms. 
    \textbf{(Top Path)} The \textit{Discriminative Paradigm} (Standard 2020--2023) employs a two-stage process where a Perception Module (CNNs/ViTs) extracts features that are explicitly linked by a Graph Inference Module (GNNs).
    \textbf{(Bottom Path)} The \textit{Generative Paradigm} (2024+) utilizes Foundation Models (LVLMs/Diffusion) to process multimodal inputs (e.g., text, kinematics, audio) holistically.
    Both approaches converge to produce a structured \textbf{Scene Graph}, which serves as the semantic engine for downstream tasks ranging from safety monitoring to generative video synthesis.}
    \label{fig:pipeline}
\end{figure}

The SG construction pipeline (Figure \ref{fig:pipeline}) relies on a combination of sophisticated computational methods for its perception and inference stages.

\paragraph{Perception: Object Detection and Feature Extraction} The perception module's primary role is to identify potential nodes and extract their features. CNN-based object detectors (e.g., YOLOv7 \citep{Yuan.2024}) and semantic segmentation models are commonly used to localize entities like instruments and anatomy \citep{Holm.2023}. The visual features for these nodes are extracted using backbones such as ResNet or Vision Transformers, sometimes adapted for surgery (e.g., EndoViT \citep{Batic.2024}). 3D data from external views often requires point cloud processing methods like PointNet++ \citep{Ozsoy.2022}, while some real-time approaches explore efficient detector-free techniques \citep{Pang.2022}. This stage outputs node attributes (class, location, visual embeddings) that feed into the graph inference module.

\paragraph{Graph Inference and Temporal Modeling} The graph inference module reasons about relationships to build the graph structure. \textbf{Graph Neural Networks (GNNs)} are central to many approaches, refining node embeddings and predicting relationship edges by passing messages across the graph structure \citep{Islam.2020, Holm.2023}. This allows entity representations to be enriched by their local context, improving interaction classification. To handle the annotation burden, some methods use latent graph models that infer an intermediate graph implicitly \citep{Murali.2024}. Given the dynamic nature of surgery, incorporating \textbf{temporal information} is paramount. This is often achieved by creating spatio-temporal graphs with explicit temporal edges \citep{Holm.2023}, using memory mechanisms to maintain temporal consistency \citep{Ozsoy.2023}, or applying sequential models like RNNs or Transformers to sequences of frame-level graphs \citep{Lin.2022}.

\paragraph{The Rise of Transformers and Foundation Models} 
\textbf{Transformer architectures} and, more recently, large foundation models have become increasingly influential, fundamentally changing how surgical scene graphs are both constructed and utilized. On one hand, these models are now being designed for end-to-end scene graph generation directly from sensor data. This trend includes influential, attention-based architectures like Rendezvous for triplet recognition \citep{Nwoye.2022} and sophisticated transformers like S2Former-OR that fuse multimodal input for holistic graph prediction \citep{Pei.2024}. This evolution has culminated in the integration of \textbf{Large Vision-Language Models (LVLMs)}, which leverage their vast pre-trained knowledge to perform zero-shot or few-shot scene graph generation, significantly reducing the dependency on large, annotated surgical datasets \citep{Ozsoy.2024b, Schmidgall.2024}. On the other hand, scene graphs are increasingly serving as a structured semantic input to guide these powerful models in complex downstream tasks. In this role, the pre-generated SG acts as a "semantic bridge," grounding the LVLM and enabling it to perform high-level cognitive tasks like nuanced Visual Question Answering (VQA) or automated report generation with greater accuracy and contextual awareness \citep{Yuan.2024, Ozsoy.2025, Lin.2022, Lin.2024}. This dual role, as both creator and consumer of structured relational knowledge, positions these advanced models as the core engine for next-generation surgical AI.

To synthesize this methodological evolution, Table \ref{tab:tech_taxonomy} provides a condensed technical taxonomy that categorizes representative studies by their perception and inference modules.

\begin{table}[htbp]
    \centering
    \caption{\textbf{Technical Taxonomy of Representative Approaches.} A condensed summary of the computational architectures used in key studies. \textit{Stages} indicates the inference process: \textbf{1} = End-to-end generation; \textbf{2} = Separate perception/inference; \textbf{Cond.} = Graph-conditioned synthesis.}
    \label{tab:tech_taxonomy}
    \footnotesize
    \setlength{\tabcolsep}{3pt}
    \renewcommand{\arraystretch}{1.1} 
    
    \begin{tabular}{@{} 
        >{\raggedright\arraybackslash}p{0.18\textwidth} 
        >{\raggedright\arraybackslash}p{0.15\textwidth} 
        >{\raggedright\arraybackslash}p{0.14\textwidth} 
        >{\centering\arraybackslash}p{0.10\textwidth} 
        >{\raggedright\arraybackslash}p{0.37\textwidth} 
    @{}}
        \toprule
        \textbf{Study} & \textbf{Perception Module} & \textbf{Inference Module} & \textbf{Stages} & \textbf{Key Technical Contribution} \\
        \midrule
        \multicolumn{5}{l}{\textit{\textbf{GNN-Based Approaches (Early Phase)}}} \\
        \citet{Islam.2020} & ResNet-18 & GCN & 2 & First use of GNNs for tool-tissue interactions. \\
        \citet{Holm.2023} & ResNet-50 & GraphSage & 2 & Dynamic SG with temporal edge connections. \\
        \citet{Seenivasan.2022b} & ResNet-101 & GAT (Attention) & 2 & Domain adaptation via graph alignment. \\
        \addlinespace[4pt]
        
        \multicolumn{5}{l}{\textit{\textbf{Transformer-Based Architectures (Middle Phase)}}} \\
        \citet{Nwoye.2022} & ResNet-18 & Transformer & 2 & \textit{Rendezvous}: Decoupled attention for triplets. \\
        \citet{Pei.2024} & PointNet++ & Transformer & 1 & \textit{S2Former}: End-to-end 3D graph generation. \\
        \citet{Li.2024} & CNN (Lite) & Transformer & 2 & Parameter-efficient lightweight architecture. \\
        \addlinespace[4pt]
        
        \multicolumn{5}{l}{\textit{\textbf{Foundation Models \& Generative AI (Recent Phase)}}} \\
        \citet{Ozsoy.2024} & CLIP (ViT) & LVLM & 1 & \textit{ORacle}: Zero-shot generation via prompting. \\
        \citet{Yuan.2024} & YOLOv7 & Cross-Attn. & 2 & \textit{SSG-VQA}: SG features query LLM decoder. \\
        \citet{Frisch.2025} & VQ-GAN & Diffusion & Cond. & \textit{SurGrID}: SG-conditioned image synthesis. \\
        \bottomrule
    \end{tabular}
\end{table}

\section{Applications and Evolution of Scene Graphs in Surgery} 
\label{sec:applications_evolution}

This chapter traces the development of surgical scene graph applications, following a narrative arc from early explorations focusing on the internal surgical field to broader OR modeling, and culminating in recent trends integrating multimodal data, language models, and emerging perspectives like egocentric vision.

\subsection{Early Focus: Internal Scene Graphs and Fine-Grained Interaction}
\label{sec:internal_sgg}

The application of scene graphs (SGs) in surgery began with the compelling need to understand the intricate web of interactions within the operative field. Early efforts logically focused on the internal view from endoscopic or laparoscopic cameras, aiming to decode the crucial interplay between instruments and anatomy.

\paragraph{Foundational Steps: Modeling Tool-Tissue Interactions}
A pivotal first step was taken by \citet{Islam.2020}, who demonstrated that a graph neural network could effectively model tool-tissue interactions from 2D endoscopic images using newly annotated data from the EndoVis 2018 challenge \citep{Allan.2020}. This work established the feasibility of learning explicit spatial and functional relationships using graph structures, laying the groundwork for subsequent research.

\paragraph{The Rise of Action Triplet Recognition}
A highly influential direction quickly emerged that focused on the specific task of recognizing surgical action triplets, structured as \texttt{<instrument, verb, target>}. \citet{Nwoye.2020} pioneered this task by proposing the first deep learning model (Tripnet) to recognize these triplets from video frames, providing a granular understanding of surgical actions. Critically, they also introduced the CholecT40 dataset, which provided the necessary data to train and evaluate such models.

The availability of this structured task and its associated datasets, particularly the expanded CholecT45/T50 benchmarks \citep{Nwoye.2022} and public challenges \citep{Nwoye.2023, Nwoye.2023b}, catalyzed rapid methodological innovation. Early models often employed attention mechanisms to pinpoint salient interactions \citep{Li.2022}, while the introduction of transformer-based architectures like the influential Rendezvous (RDV) model marked a notable advance in correlating instruments with their targets in complex scenes \citep{Nwoye.2022}.

\paragraph{Addressing Practical Hurdles in Triplet Recognition}
The computational demands of early models highlighted practical barriers to clinical use, prompting research into more efficient solutions. This led to the development of lightweight model variants \citep{Ha.2024}, detector-free approaches bypassing costly object detection steps \citep{Pang.2022}, and parameter-efficient network designs \citep{Li.2024b}. Concurrently, to mitigate the data annotation bottleneck, researchers explored techniques like mixed supervision strategies leveraging unlabeled data \citep{Sharma.2023, Sharma.2023b} and knowledge distillation methods to handle class imbalances \citep{Gui.2024}.

\paragraph{Expanding the Methodological Toolkit}
The field continued to explore diverse methods for triplet recognition. Temporal modeling became increasingly important, leading to architectures designed to capture the dynamics of surgical actions \citep{Sharma.2023, Wang.2023, Zou.2023}. Researchers also investigated alternative graph structures like Forest GCNs \citep{Xi.2022} and novel learning paradigms such as diffusion models \citep{Liu.2024}. The sophistication of these models has grown, with recent architectures like ITG-Trip incorporating temporal models such as Mamba alongside a Visual-Linguistic Association module. This module uses pre-trained text encoders to distill prior knowledge, significantly improving the recognition of rare-class interactions by bridging the gap between visual features and textual semantics \citep{Pei.2025}. Multi-task learning frameworks also demonstrated that jointly predicting triplet components or combining interaction recognition with related tasks like segmentation could improve overall performance and robustness \citep{Li.2023, Seenivasan.2022, Seenivasan.2023}.

\paragraph{Leveraging Relational Insights for Higher-Level Tasks}
Recognizing that surgical understanding requires more than identifying atomic interactions, researchers began applying graph-based representations to higher-level tasks.
\begin{itemize}
    \item \textbf{Safety Assessment:} The explicit modeling of tool-anatomy relationships proved suitable for automated safety monitoring. This includes integrating domain knowledge like Critical View of Safety (CVS) criteria into the graph structure \citep{Ban.2024} or using latent graph models to infer safety conditions with reduced annotation reliance \citep{Murali.2024}. This extends to robustly tracking critical anatomical landmarks even under severe visual occlusion. Recent knowledge-driven frameworks model the spatio-temporal relationships between tools, observable anatomy, and the entire landmark area, using the graph to reconstruct and track landmarks through complex tool-anatomy interactions \citep{Zhang.2025}.
    \item \textbf{Workflow Analysis:} Dynamic scene graphs offer a granular, explicit representation of procedural flow, which has been shown to improve accuracy and interpretability in surgical phase recognition tasks compared to methods lacking explicit interaction modeling \citep{Holm.2023, Koksal.2024}. This builds upon a rich field of workflow analysis previously surveyed by \citet{Garrow.2021}.
    \item \textbf{Generative Surgical Simulation:} In a paradigm shift from analysis to synthesis, scene graphs are now being used to control generative models. The VISAGE framework, for instance, leverages action graph triplets (\texttt{<instrument, verb, target>}) to condition diffusion models, enabling the generation of realistic future video frames of a surgical action from just a single starting frame. This opens novel avenues for data augmentation, surgical training, and simulation \citep{Yeganeh.2025}.
    \item \textbf{Related Applications:} The semantic context from SGs has also been used to enhance feature matching for challenging endoscopic 3D reconstruction tasks \citep{Yang.2022}.
\end{itemize}
This initial phase of research solidified the paradigm's relevance by concentrating on the internal view, demonstrating how the explicit relational structure of SGs could be harnessed for higher-level reasoning and laying the groundwork for broader applications.

\subsection{Expanding the View: External Scene Graphs and Holistic OR Modeling}
\label{sec:external_sgg}

While internal scene graphs detail the operative site, they omit the broader context of team dynamics and equipment status within the OR. To capture this complete picture, research expanded to encompass the entire OR environment through the development of \textbf{external scene graphs}.

\paragraph{Motivation and Data Enablement}
The goal of external OR modeling is to capture the interplay between staff, equipment, and the patient to analyze team coordination, monitor safety hazards, and optimize workflow. A major obstacle was the lack of suitable data, a gap addressed by the release of datasets like \textbf{4D-OR} \citep{Ozsoy.2022}. This benchmark provided multi-view RGB-D recordings of simulated procedures annotated with 4D semantic scene graphs, enabling the development of methods for external OR understanding.

\paragraph{Addressing Spatio-Temporal Complexity}
The dynamic external OR, with multiple moving actors and objects, requires methods that can robustly track entities and relationships in 3D space over time. Methodological advancements focused on this challenge, with models like \textbf{LABRAD-OR} incorporating memory from past graph states to improve temporal coherence and accuracy in busy scenes \citep{Ozsoy.2023}.

\paragraph{Efficient Multimodal Fusion and Knowledge Integration}
Processing multi-view and multimodal data streams efficiently became another key research focus. Architectures like \textbf{S2Former-OR} proposed single-stage transformers to integrate 2D images and 3D point clouds for direct relationship prediction \citep{Pei.2024}. The semantic complexity of the external OR also motivated the integration of domain knowledge, with frameworks incorporating surgical ontologies or predefined rules to enhance the consistency of generated graphs \citep{Dsouza.2023}. This work advanced the vision of a \textbf{Multimodal Semantic Scene Graph (MSSG)}\citep{Ozsoy.2021}, a concept now being put into practice. For example, \citet{Wagner.2024} have developed holistic, multimodal graph-based approaches that fuse imaging and non-imaging data (e.g., medical device outputs) into a unified knowledge graph to perform downstream tasks like surgical instrument anticipation. This focus on multimodality is also critical for enhancing clinical viability; frameworks like the one proposed by \citet{Bai.2025} integrate vision and kinematic data with adversarial learning to achieve robust workflow recognition even when one modality is corrupted, directly addressing the challenges of domain shifts and data noise inherent in the real-world OR \citep{Bai.2025}. These works mark significant progress in tackling the complexities of the external OR, while the persistent challenges of computational cost, occlusions, and privacy continue to drive further innovation in the field.

\subsection{Convergence: From Analysis to Intelligent Systems}
\label{sec:convergence}

The most recent phase of surgical SG research is characterized by a powerful convergence with cutting-edge AI, pushing the field from analysis towards more generalizable and intelligent systems. This evolution is defined by three interconnected themes: the rise of specialized foundation models, the use of SGs for generative AI, and a deep focus on multimodal robustness.

\paragraph{The Rise of Specialized Surgical Foundation Models}
The advent of powerful, pre-trained LVLMs has culminated in the development of foundation models specifically tailored for the surgical domain. This marks a pivotal shift from adapting generalist models, which demonstrated improved zero-shot SGG with models like \textbf{ORacle} \citep{Ozsoy.2024b}, to building holistic, multimodal systems from the ground up. The quintessential example is \textbf{ORQA}, a foundation model that unifies visual (RGB-D, egocentric), auditory, and structured data into a cohesive question-answering framework \citep{Ozsoy.2025b}. Crucially, this work demonstrates that domain-specialized models vastly outperform powerful generalist models like GPT-4 and Gemini, which struggle to interpret the complex, multimodal nature of surgical scenes. This establishes the scene graph as an essential structured reasoning component for next-generation surgical AI, powering language-based interactions like VQA \citep{Yuan.2024} and boosting performance via large-scale pre-training on surgical data like \textbf{SurgVLP} \citep{Yuan.2023}.

\paragraph{Generative AI and Controllable Simulation}
A significant new application domain has emerged where scene graphs are used to control generative AI. Moving beyond analysis, models like \textbf{SurGrID} and \textbf{VISAGE} leverage scene graphs and action triplets to condition diffusion models, enabling the high-fidelity, controllable synthesis of surgical images and videos \citep{Frisch.2025, Yeganeh.2025}. By interactively modifying the scene graph, users can precisely control the generated content, offering unprecedented potential for realistic surgical training simulators and data augmentation.

\paragraph{Multimodality and Egocentric Perspectives as a Standard}
The field is now embedding multimodality and multi-perspective views as core principles for achieving robustness. The development of comprehensive, synchronized multimodal datasets like \textbf{MM-OR} is pivotal for training these next-generation models \citep{Ozsoy.2025}. This fusion of complementary information, such as vision with kinematics, as explored by \citet{Bai.2025} through adversarial learning, produces richer context awareness and more resilient systems that can handle corrupted data or domain shifts. This holistic approach, integrating the surgeon's first-person perspective \citep{Rodin.2024, Fujii.2024} with external views, brings the field closer to realizing the vision of comprehensive representations like the MSSG \citep{Ozsoy.2021}.

\paragraph{Distinction from General Surgical LLMs}
It is important to situate these developments within the broader landscape of surgical AI. While recent works like Surgical-Mamba or Surgery-R1 utilize Large Language Models (LLMs) for surgical QA and reasoning, these models typically operate on global image tokens or pure text. This review distinguishes itself by focusing on \textit{Scene Graphs}, which provide the \textbf{explicit, structured, and spatially-grounded intermediate representation} that pure LLMs often lack. While generalist LLMs offer broad reasoning, Scene Graphs offer the grounded precision required for safety-critical tasks like robotic control, where hallucination is not an option.

\section{Results: Answering the Research Questions}
\label{sec:results}

This section synthesizes the findings from the 52 included studies to directly address the review's research questions. The analysis reveals a field of rapid growth, with distinct patterns in applications, methodologies, and data utilization. A comprehensive, filterable summary of each included study is provided in \ref{sec:appendix_tables}. An interactive, online version of these summary tables is also available on the project's Open Science Framework (OSF) repository: osf.io/jh68z.

\subsection{RQ1: Application Landscape---From Internal Tasks to Holistic Understanding}
\label{sec:application_landscape}

To address our first research question (RQ1) concerning the surgical domains and problems being solved, our analysis reveals a clear bifurcation in research focus (Table \ref{tab:app_char_matrix_final}).

The most significant cluster of research (\textbf{22 studies, 42.3\%}) concentrates on fine-grained \textbf{action triplet recognition} in the \textbf{internal surgical view}. The introduction of the CholecT50 benchmark \citep{Nwoye.2022} has acted as a major catalyst, establishing a standardized task that spurred a wave of methodological innovation in recognizing \texttt{<instrument, verb, target>} interactions. Building on this core task, other internal-view applications have emerged, including workflow analysis \citep{Holm.2023}, automated safety assessment \citep{Murali.2024, Ban.2024}, and report generation \citep{Lin.2022}.

In parallel, a distinct and growing body of work focuses on the \textbf{external operating room view} (\textbf{10 studies, 19.2\%}). These studies aim for a more holistic understanding of the OR environment by modeling interactions between staff, equipment, and the patient. Enabled by the release of simulated 4D datasets like 4D-OR \citep{Ozsoy.2022}, this research pushes towards full \textbf{4D Scene Graph Generation} to analyze team coordination and room state \citep{Ozsoy.2024, Pei.2024}.

Emerging applications are beginning to bridge these views, using the structured output of SGs to power higher-level cognitive tasks such as Visual Question Answering (VQA) \citep{Yuan.2024, Ozsoy.2024b} and, most recently, to control \textbf{generative AI} for surgical simulation \citep{Frisch.2025, Yeganeh.2025}.

\begin{table}[htbp]
    \centering
    \caption{\textbf{Frequency Count of Reviewed Studies (N=52) by Primary Application and Core Characteristics.} Categories reflect the primary focus of the study. \textit{Triplet Recognition} focuses on atomic interactions in 2D video. \textit{Generative AI} includes image/video synthesis conditioned on graphs. \textit{Foundation Models} refers to large-scale pre-trained models for holistic understanding.}
    \label{tab:app_char_matrix_final}
    \footnotesize
    \renewcommand{\arraystretch}{1.2}
    \setlength{\tabcolsep}{6pt}

    \begin{tabular*}{\textwidth}{@{\extracolsep{\fill}} l c c c c c c}
        \toprule
        & \multicolumn{2}{c}{\textbf{Static}} & \multicolumn{2}{c}{\textbf{Dynamic}} & & \\
        \cmidrule(r){2-3} \cmidrule(lr){4-5}
        \textbf{Primary Application} & \textbf{2D} & \textbf{3D} & \textbf{2D} & \textbf{4D} & \textbf{Found. Model} & \textbf{N} \\
        \midrule
        Triplet Recognition          & 8  & 0 & 14 & 0 & 0 & \textbf{22} \\
        Scene Graph Generation       & 1  & 0 & 0  & 9 & 0 & \textbf{10} \\
        Workflow Recognition         & 0  & 0 & 5  & 0 & 0 & \textbf{5}  \\
        Safety Assessment            & 1  & 0 & 3  & 0 & 0 & \textbf{4}  \\
        Generative AI                & 1  & 0 & 3  & 0 & 0 & \textbf{4}  \\
        Report Generation            & 1  & 0 & 2  & 0 & 0 & \textbf{3}  \\
        Visual Question Answering (VQA) & 2  & 0 & 0  & 0 & 0 & \textbf{2}  \\
        Foundation Model Development & 0  & 0 & 0  & 0 & 1 & \textbf{1}  \\
        Referring Inst. Segmentation & 0  & 0 & 1  & 0 & 0 & \textbf{1}  \\
        \midrule
        \textbf{Total Uses} & \textbf{14} & \textbf{0} & \textbf{28} & \textbf{9} & \textbf{1} & \textbf{52} \\
        \bottomrule
    \end{tabular*}
\end{table}

\subsection{RQ2: Methodological and Data Trends---The Path to Foundation Models}
\label{sec:methodological_and_data_trends}

Addressing RQ2, we identified a clear methodological trajectory and a critical divide in data utilization.

\begin{figure}[htbp]
    \centering
    \includegraphics[width=\textwidth]{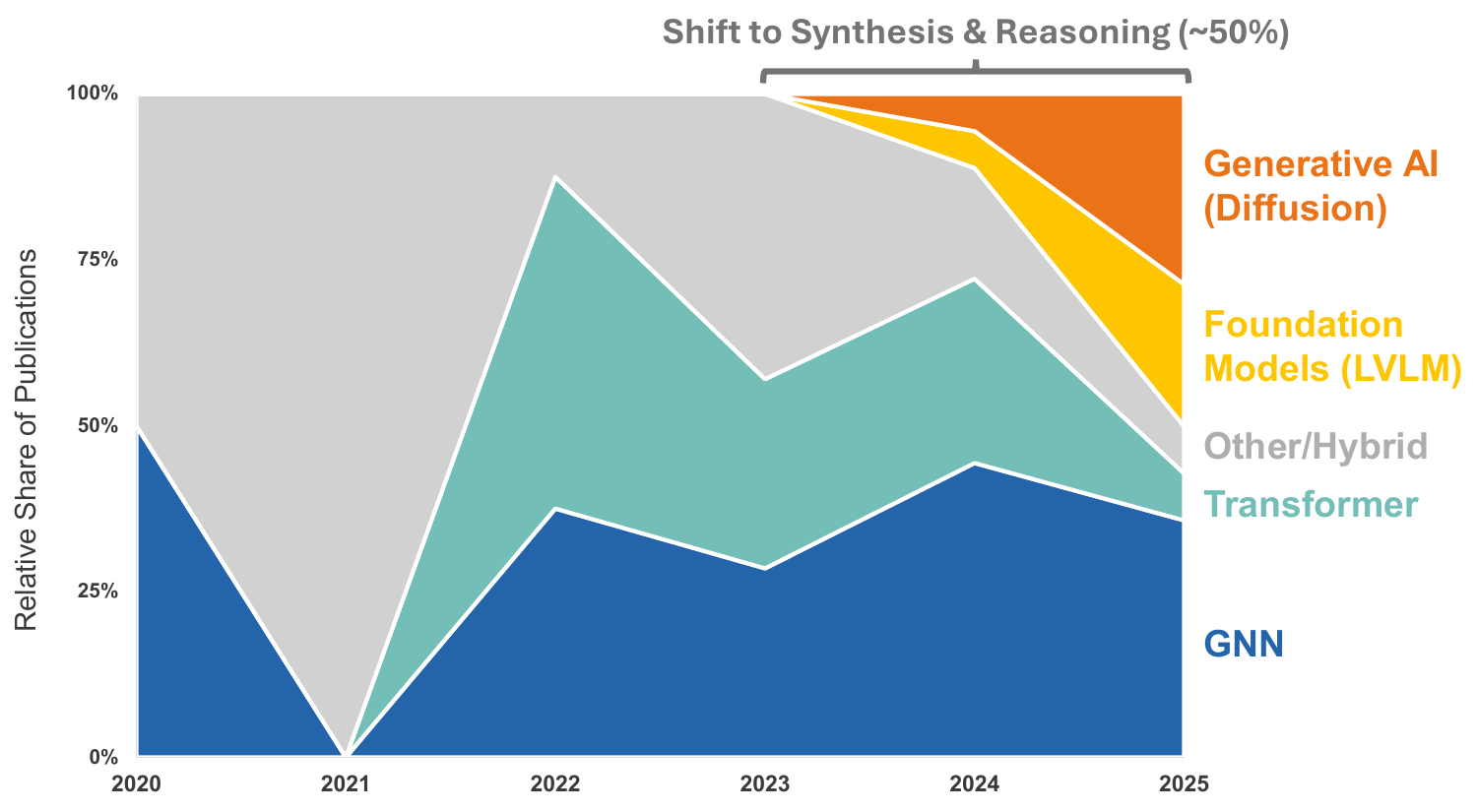}
    \caption{\textbf{The Methodological Evolution of Surgical Scene Graph Research.} This 100\% stacked area chart illustrates the shifting distribution of core computational methods over time ($N=52$). While the field was initially established on discriminative \textbf{GNN} (blue) and \textbf{Transformer} (teal) architectures, the modern era (2024--2025) marks a decisive paradigm shift. The rapid emergence of \textbf{Foundation Models (LVLMs)} for reasoning and \textbf{Generative AI} for simulation now accounts for approximately 50\% of recent contributions, highlighting the field's trajectory from pure scene analysis to synthesis and actionable intelligence.}
    \label{fig:method_evolution_viz}
\end{figure}

\paragraph{The Methodological Trajectory: From GNNs to Foundation Models} The computational methods have matured significantly, mirroring broader trends in AI (Figure \ref{fig:method_evolution_viz}).
\begin{itemize}
    \item \textbf{Foundational Approaches:} Early pioneering work established the viability of using \textbf{Graph Neural Networks (GNNs)} to explicitly model tool-tissue relationships and enrich node features with contextual information \citep{Islam.2020}.
    \item \textbf{The Rise of the Transformer:} The field saw a decisive shift towards \textbf{Transformer-based architectures}, whose attention mechanisms proved highly effective at correlating entities in complex scenes. The influential Rendezvous model, for example, became a standard for the triplet recognition task \citep{Nwoye.2022}, and sophisticated Transformers are now being used for end-to-end multimodal SGG from 4D data \citep{Pei.2024}.
    \item \textbf{The Dawn of Foundation Models:} The most recent trend is the integration of \textbf{Large Vision-Language Models (LVLMs)}. These models are used in two main ways: either for direct zero-shot scene graph generation, leveraging their vast pre-trained knowledge \citep{Ozsoy.2024b}, or by using a generated SG as a structured semantic input to guide the LVLM in complex reasoning tasks like VQA \citep{Yuan.2024}. This trend culminates in domain-specific foundation models like ORQA, which are shown to outperform general-purpose models on surgical tasks \citep{Ozsoy.2025b}. However, this shift introduces a new tension between semantic complexity and verifiability. While generalist LVLMs offer broad reasoning capabilities, they lack the grounded precision of earlier discriminative GNNs. The integration of SGs into these architectures represents a strategic return to interpretability, acting as a constraint mechanism to ground the probabilistic outputs of foundation models in the physical reality of the OR.
\end{itemize}

Our quantitative analysis reveals a decisive methodological shift. In the early era ($\le$2022), architectures were evenly split between GNNs and Transformers. However, in the modern era ($\ge$2023), specialized Foundation Models and Generative AI (Diffusion) have surged. By 2025, these two categories combined account for approximately \textbf{50\%} of new research output, underscoring the field's rapid pivot from analysis to synthesis.

\begin{figure}[htbp]
    \centering
    \includegraphics[width=\textwidth]{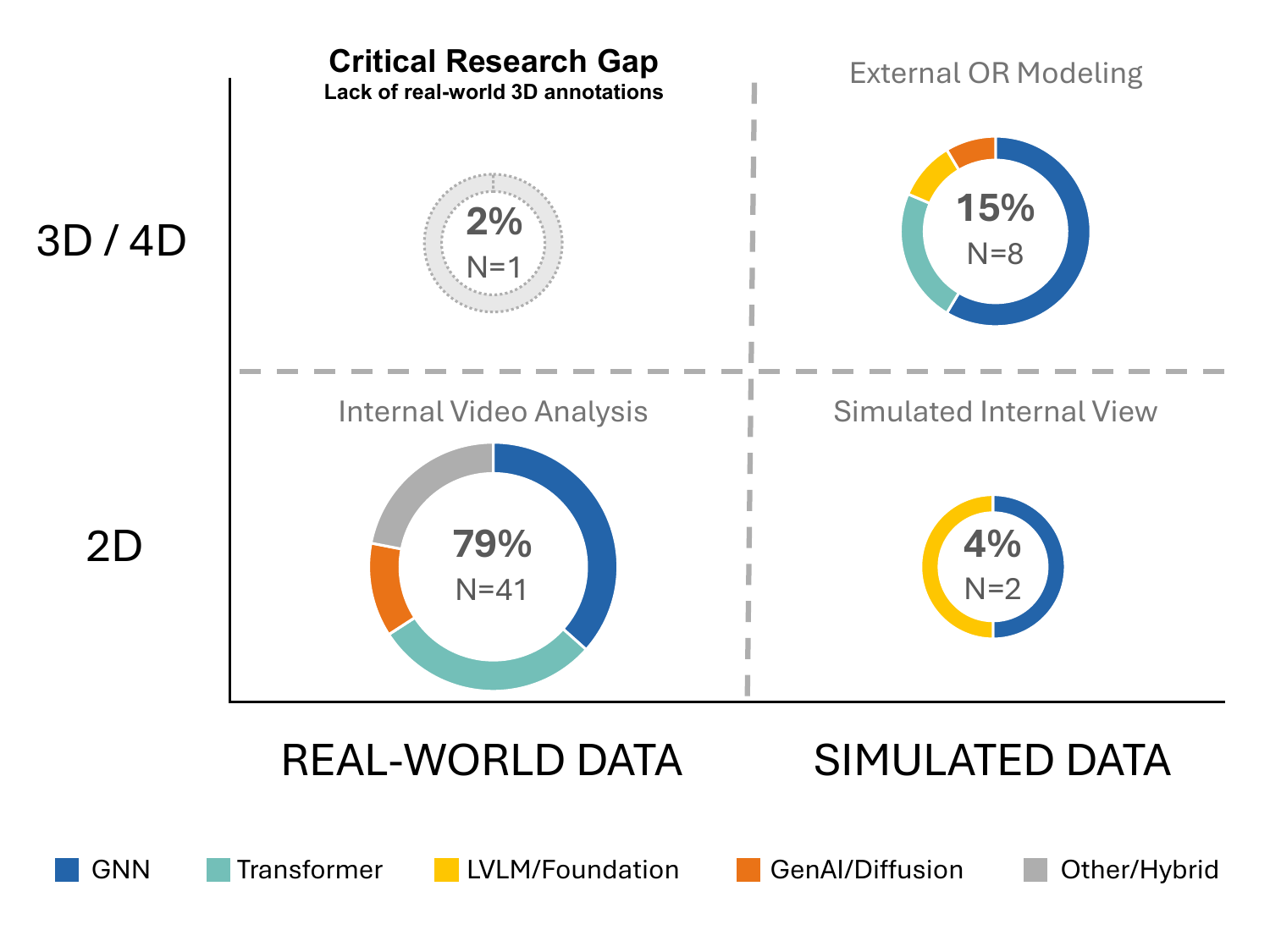}
    \caption{\textbf{Quantifying the Data Divide in Surgical Scene Graph Research.} This $2 \times 2$ matrix categorizes the 52 included studies based on Data Source (Real vs. Simulated) and Dimensionality (2D vs. 3D/4D). Donut charts are scaled to represent the volume of research ($N$), with colored slices indicating the method distribution (following the color legend of Fig.~\ref{fig:method_evolution_viz}). The analysis reveals a heavy concentration of research in \textbf{Internal Video Analysis} (Real/2D, 79\%), while \textbf{External OR Modeling} (15\%) remains largely dependent on simulation. The ``ghost'' donut in the top-left quadrant highlights the \textbf{Critical Research Gap}: the lack of dynamic, multi-sensor 3D scene graph models validated on real-world clinical data.} 
    \label{fig:data_divide_plot}
\end{figure}

\paragraph{The Data Divide: Real-World 2D Video vs. Simulated 4D Worlds}
Our analysis of data sources and modalities uncovers a crucial 'data divide' (Figure \ref{fig:data_divide_plot}).
\begin{itemize}
    \item \textbf{Real-World 2D Research:} This domain dominates the field, accounting for \textbf{79\% (N=41)} of all included studies. It is driven by the availability of endoscopic video datasets (e.g., CholecT50) and focuses primarily on internal interaction recognition using GNNs and Transformers.
    
    \item \textbf{Simulated 4D Modeling:} In sharp contrast, external OR modeling relies heavily on simulated environments. The top-right quadrant of Figure \ref{fig:data_divide_plot} shows that \textbf{15\% (N=8)} of studies operate here, utilizing the ground-truth 3D positions provided by datasets like 4D-OR.
    
    \item \textbf{The Critical Research Gap:} The top-left quadrant exposes the field's primary bottleneck. Only one study ($N=1$, \citep{Ozsoy.2021}) appears in the Real-World 3D/4D domain, and it represents a conceptual framework on sparse data rather than a deployed system. There are currently \textbf{zero} studies performing fully dynamic, external 3D/4D scene graph generation on authentic, multi-sensor data from a real operating room.
\end{itemize}

This distribution is not coincidental but reflects a trade-off between \textbf{Semantic Complexity} and \textbf{Data Reality}. The top-left quadrant represents the "Holy Grail" of surgical computer vision---dynamic, 4D understanding of real-world scenes---which remains sparsely populated due to the exponential difficulty of annotating 4D graphs on authentic, occluded, and messy surgical video compared to 2D bounding boxes or simulated environments.

\subsection{Synthesizing the Benefits and Gaps (RQ3 \& RQ4)}
\label{sec:benefits_and_gaps}

Finally, by synthesizing the findings related to applications (RQ1) and methods (RQ2), we can identify the primary benefits and challenges (RQ3) and set the stage for future research directions (RQ4).

The core benefit of SGs lies in their ability to provide an explicit, relational structure that powers higher-level reasoning. The reviewed literature offers robust quantitative evidence that explicit relational modeling consistently outperforms unstructured baselines. To illustrate this impact, Table~\ref{tab:performance_gains} highlights representative performance gains across three evolutionary stages of surgical AI applications.

\begin{table*}[htbp]
    \centering
    \caption{\textbf{Representative Performance Gains by Task.} This table highlights the quantifiable benefits of Scene Graph models compared to baselines across three core evolutionary stages of surgical AI. Note: `Abs.' denotes an absolute percentage point or metric improvement.}
    \label{tab:performance_gains}
    \footnotesize 
    \renewcommand{\arraystretch}{1.3} 
    \begin{tabular*}{\textwidth}{@{\extracolsep{\fill}} l l l l c c @{}}
        \toprule
        \textbf{Study} & \textbf{Dataset} & \textbf{Baseline} & \textbf{Metric} & \textbf{Gain} & \textbf{Type} \\
        \midrule
        \multicolumn{6}{@{}l}{\textbf{Action Triplet Recognition}} \\
        \citet{Nwoye.2022} & CholecT50 & TripNet & mAP & +9.9\% & Abs. \\
        \citet{Shen.2025} & EndoVis2018 & MSLRG & mAP & +10.14\% & Abs. \\
        \addlinespace
        \multicolumn{6}{@{}l}{\textbf{Workflow \& Safety Assessment}} \\
        \citet{Holm.2023} & CATARACTS & Static Graph & Macro F1 & +14.66\% & Abs. \\
        \citet{Ban.2024} & CVS100 & TSM & Accuracy & +9.4\% & Abs. \\
        \addlinespace
        \multicolumn{6}{@{}l}{\textbf{Foundation Models \& Generative AI}} \\
        \citet{Ozsoy.2025c} & MM-OR & ORacle & Macro F1 & +6.2\% & Abs. \\
        \citet{Yeganeh.2025} & CholecT50 & SVD Baseline & FVD ($\downarrow$) & -2090 & Abs. \\
        \bottomrule
    \end{tabular*}
\end{table*}

To contextualize the impact of these methods, we summarize the mechanistic advantages driving these advancements across the field's evolutionary trajectory:

\paragraph{Atomic Understanding (Triplet Recognition)} In highly structured tasks, explicitly modeling dependencies overcomes the limitations of isolated object detectors, which struggle with occlusions and instrument-tissue overlaps. By enforcing relational priors, the graph-based \textit{Rendezvous} model achieved a +9.9\% absolute increase in mean Average Precision (mAP) over an unstructured baseline \citep{Nwoye.2022}. Recent multimodal graph networks have extended these gains to +10.14\% mAP by leveraging cross-modal associations \citep{Shen.2025}.

\paragraph{Contextual Reasoning (Workflow \& Safety)} Frame-by-frame classifiers inherently lack the temporal and spatial awareness required for complex procedural reasoning. Incorporating dynamic SGs provides this missing context. For example, explicitly adding temporal edges to transition from a static to a dynamic scene graph improved surgical workflow recognition by an absolute +14.66\% in Macro F1 \citep{Holm.2023}. Similarly, explicitly grounding safety criteria through concept graph networks improved Critical View of Safety (CVS) assessment by +9.4\% in average accuracy over temporal shift modules (TSM) \citep{Ban.2024}, as the graph structure actively prevents the network from predicting safety if prerequisite anatomical relationships are missing.

\paragraph{Next-Generation Systems (Foundation Models \& Generative AI)} In the realm of complex reasoning, SGs act as essential neuro-symbolic anchors. The \textit{MM2SG} model---the first multimodal large language model for scene graph generation---outperformed the \textit{ORacle} baseline by an absolute +6.2\% in Macro F1 \citep{Ozsoy.2025c}, demonstrating that integrating modalities like audio and robotic logs into the graph yields a more holistic understanding of the scene. Furthermore, in novel video synthesis tasks, conditioning diffusion models on action graphs considerably reduces frame hallucinations, demonstrating an absolute reduction of 2090 points in Fréchet Video Distance (FVD) compared to standard baselines \citep{Yeganeh.2025}.

\paragraph{A Caveat on Aggregate Metrics} While these consistent performance gains validate the hypothesis that imposing graph structure is crucial for decoding surgical complexity, it is important to view these improvements critically. As we will argue in Section \ref{sec:limitations}, aggregate metrics such as mAP and F1-score provide foundational utility but can obscure safety-critical edge cases, underscoring the need for more clinically aligned evaluation frameworks. Furthermore, this success is currently confined to narrow methodological pockets. The ``data divide'' identified in Section \ref{sec:methodological_and_data_trends} reveals that while algorithms are maturing, their application is restricted by distinct research ``coldspots'':

\begin{itemize}
    \item \textbf{The Real-World Validation Gap:} As quantified above, the field lacks studies that validate holistic, external 4D SGG models on complex, real-world clinical data. Bridging the gap between simulated 4D training and real-world deployment is the primary translational barrier.
    \item \textbf{Limited Procedural Diversity:} Quantitative analysis reveals that Laparoscopic Cholecystectomy dominates the landscape, accounting for 44.2\% (23/52) of included studies, largely driven by the CholecT40/T50 benchmarks. Orthopedic Surgery (primarily simulated) follows at 15.4\% (8/52), while Cataract Surgery accounts for 11.5\% (6/52). This concentration creates significant "coldspots" in other complex domains like neurosurgery or open surgery.
    \item \textbf{Underutilization of Multimodality:} While emerging, the fusion of SGs with non-visual data streams like device kinematics \citep{Bai.2025}, audio, or EHRs remains a significant area for growth.
\end{itemize}

\section{Discussion: Gaps, Challenges, and Future Directions}
\label{sec:discussion}

The journey of scene graphs into surgery marks a significant step towards deeper, contextual understanding of complex medical procedures. As synthesized in the preceding sections, surgical scene graphs have demonstrated their potential across a spectrum of applications, from fine-grained interaction recognition \citep{Nwoye.2022} to holistic OR modeling \citep{Ozsoy.2024b}. Yet, despite this rapid evolution, the field is now poised to tackle a new set of challenges on the path to clinical translation. This chapter discusses the remaining limitations, outlines the key future directions that will shape the next wave of innovation, and reflects on the practical implications for surgery.

\subsection{Current Limitations and Consolidated Challenges}
\label{sec:limitations}

\begin{figure}[htbp]
    \centering
    \includegraphics[width=\textwidth]{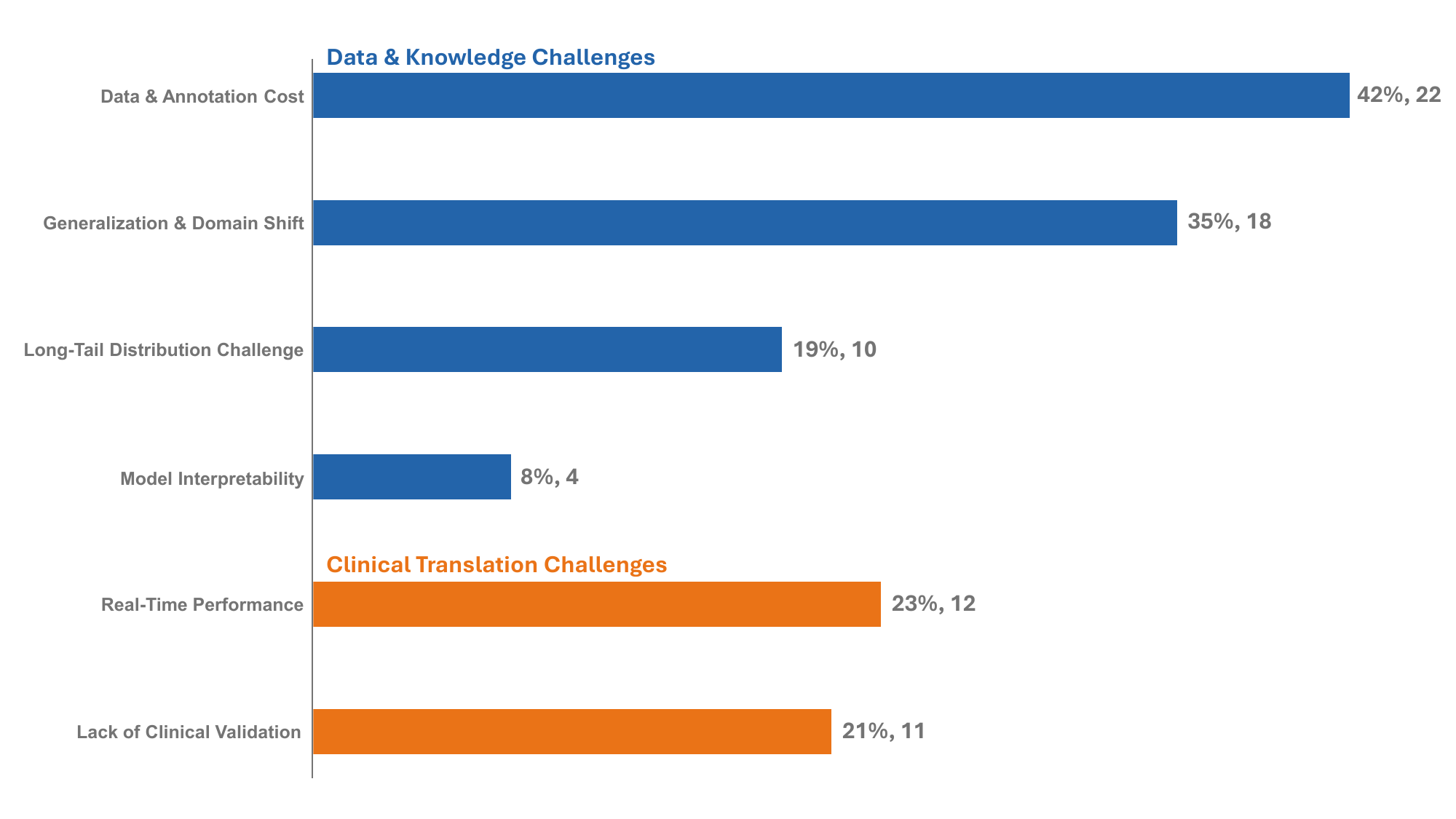}
    \caption{\textbf{Frequency of Cited Limitations in the Reviewed Literature ($N=52$).} This chart synthesizes the challenges most frequently identified by the authors, categorized into \textbf{Data \& Knowledge} constraints and \textbf{Clinical Translation} barriers. While 21\% of papers explicitly cite a \textbf{Lack of Clinical Validation} as a limitation, our methodological analysis reveals a deeper gap: \textbf{100\%} of the included studies rely on retrospective offline validation, with zero systems currently deployed in live clinical workflows.}
    \label{fig:limitations}
\end{figure}

\paragraph{Data Scarcity and Annotation Bottleneck} Perhaps the most pervasive challenge is the 'Achilles' heel' of data. Creating large-scale, diverse datasets is exceptionally demanding, as scene graph annotation requires expert clinical knowledge to label not only instruments and fine-grained anatomy but also the complex, often subtle, relationships between them \citep{Islam.2020, Yuan.2024}. This expensive and time-consuming process limits the size and procedural variety of available datasets. This scarcity directly contributes to two other major issues: poor model generalization and difficulties in handling the \textbf{long-tail distribution} of surgical events, where rare but critical interactions or complications are vastly underrepresented, making them difficult for models to learn \citep{Koksal.2024}. While mitigation strategies like foundation model pre-training \citep{Ozsoy.2025b}, weak supervision \citep{Sharma.2023}, and latent representations \citep{Murali.2024} are being explored, the fundamental data bottleneck persists. To quantify these challenges, we performed a synthesis of the limitations reported within the included literature, the results of which are shown in Figure~\ref{fig:limitations}.

\paragraph{From Atomic Detection to Multi-Hop Reasoning} While the field has mastered the detection of atomic triplets (e.g., \texttt{<grasper, holds, tissue>}), there remains a limitation in structural depth. Most current approaches utilize the scene graph effectively as a list of independent detections rather than a connected web of knowledge. True graph-based reasoning---traversing multi-hop connections to understand indirect relationships (e.g., recognizing that a \texttt{retractor} pulling \texttt{tissue A} is exerting tension on \texttt{artery B})---remains largely unexplored. Moving from simple triplet detection to complex query reasoning is essential to unlock the full semantic potential of the graph structure.

\paragraph{Technical Hurdles: Performance, Robustness, and Generalization} A primary technical barrier is achieving the performance and reliability required for clinical use. Many envisioned intra-operative applications, such as real-time alerts, demand low-latency processing. While research into efficient architectures and model distillation is promising \citep{Ozsoy.2025b, Ha.2024, Li.2024b}, real-time execution for complex scene analysis remains a significant engineering challenge. Performance is further compromised by challenging visual conditions, and inaccuracies can propagate to undermine the reliability of \textbf{downstream applications} \citep{Murali.2024}. Finally, despite advances with foundation models, ensuring robust \textbf{generalization and domain shift} across different surgeons, patients, and equipment remains a critical barrier to widespread clinical translation \citep{Bai.2025, Ozsoy.2024b}.

\paragraph{Ethical Considerations and Clinical Trust} The detailed analysis of OR activities raises significant ethical questions. Continuously monitoring interactions involving patients and staff requires careful attention to privacy, data security, and informed consent. Furthermore, potential biases within training datasets could lead to performance disparities. Ensuring the safety, reliability, and \textbf{interpretability} of these complex models is paramount for gaining clinical acceptance and trust. Addressing these ethical dimensions proactively is a crucial prerequisite for responsible integration.

\paragraph{The Clinical Translation and Evidence Gap} 
A critical bottleneck identified by our systematic mapping is the complete absence of prospective clinical validation. While \textbf{80.8\%} (42/52) of included studies utilize real-world surgical data, \textbf{100\%} of these efforts remain retrospective analyses of offline datasets (e.g., Cholec80, CATARACTS). Furthermore, a significant disconnect exists between technical metrics and clinical safety. The vast majority of studies rely on mean Average Precision (mAP) or F1-score. While standard in computer vision, these aggregate metrics mask safety-critical failures; a model can achieve a high mAP while missing a single, catastrophic interaction (e.g., \texttt{<grasper, injuring, artery>}) that would be unacceptable in a live scenario. Regarding experimental rigor, while \textbf{76.9\%} of studies demonstrate quantitative improvements over state-of-the-art baselines, only \textbf{15.4\%} (8/52) perform rigorous ablation studies to isolate the specific contribution of the relational modeling.

\subsection{Position within the Surgical Foundation Model Landscape}
It is important to situate Scene Graphs within the broader, rapidly exploding landscape of surgical Large Language Models (LLMs) and Large Vision-Language Models (LVLMs). Recent works such as Surgery-R1 \citep{Hao.2025b}, Surgical-Mamba \citep{Hao.2025}, and EndoChat \citep{Wang.2025} have demonstrated impressive capabilities in surgical question answering and reasoning. 

In contrast, the research surveyed here positions Scene Graphs not as a competitor to LLMs, but as a necessary \textbf{Neuro-Symbolic Guardrail}. We argue that the future standard for Surgical AI will not be end-to-end LVLMs, but \textbf{Graph-Grounded Agents}. In this architecture, the SG acts as a semantic constraint: the LLM may propose an action based on statistical likelihood, but the execution is gated by the presence of specific edges in the dynamic scene graph. For instance, if a generative agent proposes the action \texttt{incise cystic duct}, the architectural logic would first query the real-time Scene Graph for the precondition \texttt{<hepatocystic\_triangle, is, dissected>}. If this edge is absent, indicating the Critical View of Safety is not met, the system blocks the autonomous action and prompts the surgeon, effectively preventing the hallucinated or premature step. This distinction is paramount for \textbf{regulatory approval and clinical safety}. While a generalist LLM might hallucinate a dangerous step, a scene graph-based system provides a verifiable audit trail. As shown by \citet{Ozsoy.2025b}, specialized foundation models that incorporate these structured representations significantly outperform generalist models (like GPT-4o), validating that SGs are the essential 'anchor' required to prevent hallucination in high-stakes surgical environments.

\subsection{Future Directions: From Scene Understanding to Surgical Agency}
\label{sec:future_directions}

The trajectory of surgical SG research is now moving beyond perception and interpretation toward more integrated, intelligent, and clinically actionable systems. The next wave of innovation will likely focus on the following key areas.

\paragraph{Towards Actionable Intelligence and Surgical Agency} While foundation models can now interpret the OR with impressive accuracy \citep{Ozsoy.2025b}, the next frontier is \textbf{actionable intelligence}. Future work will focus on closing the loop between the scene graph engine and physical or digital systems. This involves using the structured understanding of the SG to not just describe the scene, but to \textbf{drive context-aware robotic actions}, generate \textbf{dynamic AR guidance}, or trigger \textbf{adaptive alerts}, moving the paradigm from passive analysis to active assistance.

\paragraph{Grounding in Patient-Specific and Procedural Context} A critical next step is to ensure that AI reasoning is \textbf{grounded in the specific clinical reality} of each case. This involves developing methods to fuse real-time scene graphs with pre-operative data (e.g., patient-specific anatomical models from CT scans), surgical plans, and even surgeon-specific preferences. The goal is to create systems that provide not just generic advice, but personalized, contextually precise recommendations.

\paragraph{Incorporating Causal and Counterfactual Reasoning} Current SGs excel at modeling correlations (what is happening), but future systems will need to understand \textbf{causality} (why it is happening). Research will likely explore causal graph neural networks and counterfactual reasoning to predict the consequences of potential actions. A system that can reason about "what would happen if..." could provide invaluable decision support for avoiding complications and optimizing surgical strategy.

\paragraph{Generative Worlds for Training and Planning} The emergence of controllable simulation is a significant breakthrough \citep{Frisch.2025, Yeganeh.2025}. The future lies in creating fully interactive \textbf{"digital twin" operating rooms}. In these virtual environments, surgeons could train on rare or complex scenarios generated by an AI or even rehearse a specific patient's surgery using a simulation grounded in their pre-operative imaging, with the scene graph acting as the underlying world model.

\paragraph{The Validation and Trustworthiness Bottleneck} As these systems become more capable, the primary barrier to adoption will shift from technical feasibility to \textbf{clinical validation and trust}. The next few years must see a concerted effort to develop rigorous frameworks for evaluating the safety, reliability, and explainability of these complex models in real-world clinical settings. Moving beyond computer vision metrics to define clinically meaningful evaluation protocols will be paramount for regulatory approval and surgeon acceptance.

\subsection{Practical Implications}
\label{sec:implications}

The successful integration of surgical scene graphs holds profound practical implications for surgery and patient care. Foremost is the potential for \textbf{enhanced patient safety} through automated detection of high-risk situations and verification of safety protocols like the CVS \citep{Murali.2024, Ban.2024}. Concurrently, the structured representation of activities promises \textbf{improved surgical workflow and efficiency} by enabling objective analysis of team coordination, anticipation of instrument needs, and automated generation of detailed operative reports \citep{Holm.2023, Ozsoy.2023, Wagner.2024}.

\begin{figure}[htbp]
    \centering
    \includegraphics[width=1\textwidth]{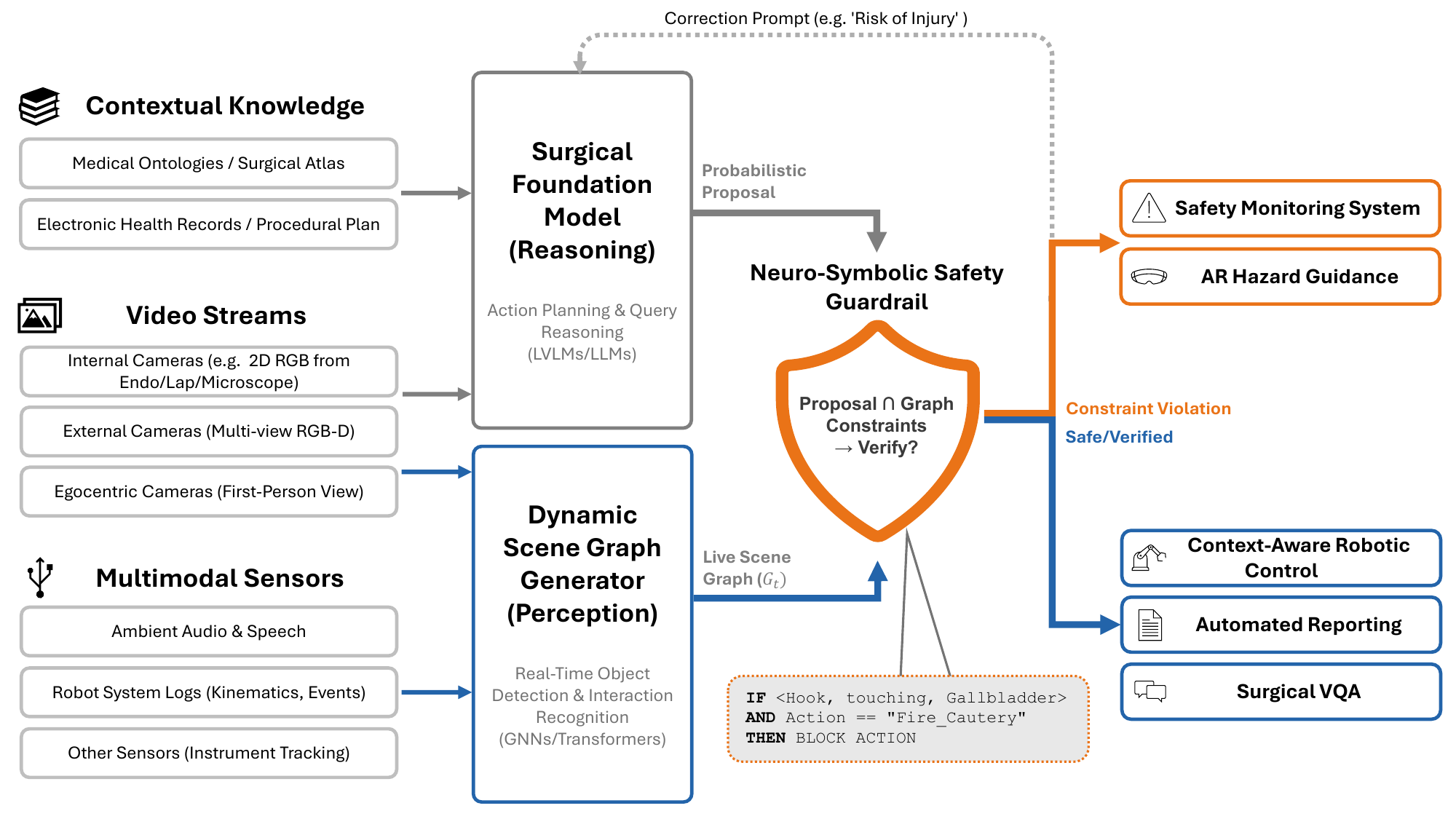} 
    \caption{\textbf{Conceptual Architecture of a Neuro-Symbolic Surgical Operating Ecosystem.} 
    Unlike purely generative approaches, this architecture introduces a \textbf{Neuro-Symbolic Guardrail} to prevent hallucinations. A \textbf{Dynamic Scene Graph Generator} (Perception) constructs a verifiable ``World Model'' ($G_t$) based on real-time sensor data. Simultaneously, a \textbf{Surgical Foundation Model} (Reasoning) generates probabilistic action plans. Crucially, the Guardrail cross-references these proposals against the structured constraints of the Scene Graph (see callout). Validated actions proceed to \textbf{Clinical Action} (e.g., Robotics, Reporting), while violations trigger \textbf{Safety Interventions} (e.g., AR Warnings) and provide a corrective feedback loop to update the model's context.}
    \label{fig:system_view}
\end{figure}

Furthermore, SGs offer a foundation for more \textbf{objective surgical skill assessment and training}, moving beyond simple metrics to allow quantitative analysis of technique and decision-making patterns, a potential amplified by emerging egocentric perspectives and generative simulation \citep{Frisch.2025}. Finally, SGs are positioned to serve as the essential semantic 'cognitive map' for \textbf{advanced surgical assistance and automation}, enabling context-aware robotic platforms and intelligent augmented reality guidance \citep{Yuan.2024}. A conceptual vision for such an integrated system is presented in Figure~\ref{fig:system_view}.

Realizing these benefits mandates strong interdisciplinary collaboration among computer scientists, engineers, surgeons, and ethicists to navigate the complex technical, validation, and translational hurdles inherent in bringing such sophisticated AI tools safely into the clinical environment.

\subsection{Limitations of this Review}
\label{sec:review_limitations}

While this scoping review was conducted according to a rigorous protocol, several limitations should be considered. First, as is characteristic of a scoping review, we did not perform a formal quality assessment (critical appraisal) of the included studies. Our primary objective was to map the breadth and evolution of the field, not to synthesize evidence based on study quality. Second, our search strategy, though comprehensive, has boundaries. By focusing on peer-reviewed literature published since 2019 in major English-language databases, we may have excluded some earlier foundational works, non-English studies, or very recent preprints. The pragmatic limit applied to the Google Scholar search could also have resulted in missed articles. Despite these constraints, this review provides the first systematic map of the surgical scene graph landscape, offering a robust foundation for future research.

\section{Conclusions}
\label{sec:conclusions}

This scoping review has charted the emergence and evolution of scene graphs as a powerful paradigm for understanding the complex, dynamic environments of surgery. Originating from computer graphics and vision, scene graphs provide structured, semantic representations of surgical entities and their interactions, facilitating a shift beyond simple object detection towards richer relational reasoning vital for advanced surgical data science.

\subsection{Key Findings Synthesized} 
\label{sec:key_takeaways}

Our synthesis reveals that the core strength of surgical scene graphs lies in their ability to uniquely encode \emph{how} entities relate, offering an interpretable foundation for sophisticated surgical AI. This application breadth has been paralleled by rapid methodological advancements, progressing from initial GNNs to the current paradigm of \textbf{specialized, multimodal foundation models}. This progress has culminated in the development of systems that can perform complex reasoning, robustly handle data corruption, and even enter the novel domain of \textbf{generative AI for surgical simulation}. Despite persistent challenges in data scarcity and real-time performance, they are now being actively addressed through transfer learning, model distillation, and adversarial frameworks. Consequently, the field's trajectory is clear: surgical scene graphs have matured from simple descriptive tools into an essential \textbf{verification layer} for modern AI. By providing the structural grounding that foundation models lack, Scene Graphs are poised to become the standard interface for safe, hallucination-free intelligent surgical systems.

\subsection{Recommendations for Future Research}
\label{sec:recommendations}

To realize the full potential of surgical scene graphs, we recommend future efforts prioritize the following key areas:

\begin{enumerate}[leftmargin=*, itemsep=2pt, topsep=3pt]
    \item \textbf{Develop Neuro-Symbolic and Closed-Loop Systems:} Shift focus from purely descriptive models to \textbf{Hybrid Neuro-Symbolic Architectures}. In this paradigm, the Scene Graph serves as the grounded ``World Model'' (long-term memory) while an LVLM acts as the ``Reasoning Agent'' (short-term planning). This separation of concerns creates a verification loop that can safely drive real-time robotic actions.
    
    \item \textbf{Focus on Clinical Grounding and Personalization:} Create architectures that can fuse real-time scene graphs with patient-specific data (e.g., pre-operative scans, EHRs) and procedural plans to provide personalized, rather than generic, guidance.
    
    \item \textbf{Explore Causal and Counterfactual Reasoning:} Move beyond correlational models to investigate causal inference, enabling systems to reason about the potential consequences of surgical actions (``what would happen if...'') to provide proactive decision support.
    
    \item \textbf{Build Interactive Generative Environments:} Leverage the power of SG-conditioned generative models to build fully interactive "digital twin" simulations for surgical training, planning, and the validation of autonomous systems.
\end{enumerate}

Finally, regarding evaluation standards, we propose a shift away from standard computer vision metrics.

\begin{tcolorbox}[
    colback=gray!5!white, 
    colframe=gray!50!black,
    title=\textbf{\textsc{Proposal for Future Benchmarking:} \\ \textsc{The Validation Trinity}},
    fonttitle=\bfseries\large,
    halign title=center, 
    toptitle=3mm,        
    bottomtitle=3mm,     
    sharp corners,
    boxrule=1pt,
    width=\textwidth,
    before=\par\vspace{1em}\noindent,
    after=\vspace{1em}   
]
To bridge the gap between technical metrics and clinical utility, we propose that future surgical scene graph contributions must report performance across three distinct axes:

\begin{description}[style=nextline, leftmargin=0pt, font=\bfseries, itemsep=8pt]
    
    \item[1. Semantic Query Success (SQS)] 
    \textit{Can the graph answer the clinical question?} \\
    Moving beyond mAP, this metric evaluates the ability to correctly resolve high-level safety queries (e.g., ``Is the Critical View of Safety criteria met?'') derived from the graph structure.
    
    \item[2. Latency-Aware Accuracy (LAA)] 
    \textit{Is the graph valid at the moment of decision?} \\
    Given the dynamic nature of surgery, accuracy must be weighted by inference speed. A scene graph that updates slower than the surgeon's cognitive reaction time ($\sim$300ms) relies on obsolete state data.
    
    \item[3. Safety-Critical Recall (SCR)] 
    \textit{Does the model catch the "black swan" events?} \\
    Aggregate metrics mask failures in rare classes. SCR specifically isolates performance on high-risk, long-tail classes (e.g., \texttt{<instrument, injuring, vessel>}) where a single false negative is unacceptable.

\end{description}
\end{tcolorbox}

In conclusion, surgical scene graphs represent a vibrant and rapidly advancing frontier in surgical data science. By continuing to innovate in methods, data, and evaluation, while maintaining a strong focus on clinical relevance and ethical considerations, the field is well-positioned to deliver technologies that fundamentally improve how we perceive, analyze, and interact within the OR, ultimately benefiting both surgeons and patients.

\section*{Funding}
This paper is supported by the state of Bavaria through Bayerische Forschungsstiftung (BFS) under Grant AZ-1592-23-ForNeRo.

\section*{Acknowledgments}
The authors would like to thank Ege Özsoy for providing valuable feedback and critical comments on the manuscript. His suggestions were instrumental in improving and refining this paper.

\section*{Declaration of Competing Interest}
Conflicts of interest: none. The authors declare no known competing financial interests or personal relationships that could have appeared to influence the work reported in this paper. For full transparency, we note that author Nassir Navab is a regular member of the editorial board of Medical Image Analysis, but was not involved in the editorial handling or peer review of this manuscript.

\section*{Data Statement}
The data for this scoping review consists of the collection of identified and analyzed peer-reviewed articles. The detailed search strategy, including databases, search strings, and inclusion/exclusion criteria, is described in the Methodology section. All articles included in the final synthesis are cited within the manuscript and listed in the References section. The data charting form is available from the corresponding author upon reasonable request. The review protocol was registered on the Open Science Framework (OSF; osf.io/jh68z), where the data charting form and an interactive version of the summary tables from Appendix A, including the code used to analyze the data, are also publicly hosted.

\section*{Declaration of Generative AI and AI-assisted technologies in the writing process}
During the preparation of this work the author(s) used Gemini in order to improve language and clarity, assist with LaTeX formatting, and help structure the manuscript according to the journal's guidelines. After using this tool, the author(s) reviewed and edited the content as needed and take full responsibility for the content of the published article.

\section*{CRediT authorship contribution statement}
\textbf{Angelo Henriques:} Conceptualization, Methodology, Investigation, Data Curation, Writing – Original Draft, Visualization. \textbf{Korab Hoxha:} Conceptualization, Investigation, Data Curation, Writing – Review \& Editing. \textbf{Daniel Zapp:} Writing – Review \& Editing, Resources. \textbf{Peter C. Issa:} Writing – Review \& Editing, Resources. \textbf{Nassir Navab:} Conceptualization, Writing – Review \& Editing, Supervision. \textbf{M. Ali Nasseri:} Conceptualization, Writing – Review \& Editing, Resources, Supervision, Funding acquisition.

\appendix
\section{Summary Tables of Included Studies}
\label{sec:appendix_tables}

Note: An interactive, filterable version of these tables is available at osf.io/jh68z.

\begingroup
\footnotesize
\setlength\LTcapwidth{\textwidth}

\begin{longtable}{@{} p{0.25\textwidth} p{0.2\textwidth} p{0.25\textwidth} p{0.12\textwidth} @{}}
    \caption{\textbf{Surgical Scene Graphs - Internal View and Related Applications.}}
    \label{tab:summary_internal_related_split} \\
    \toprule
    \textbf{Author \& Year} & \textbf{Domain} & \textbf{Application} & \textbf{SG Char.} \\
    \midrule
\endfirsthead

    \multicolumn{4}{c}%
    {{\tablename\ \thetable{} -- \textit{Continued from previous page}}} \\
    \toprule
    \textbf{Author \& Year} & \textbf{Domain} & \textbf{Application} & \textbf{SG Char.} \\
    \midrule
\endhead

    \bottomrule
    \multicolumn{4}{r}{{\textit{Continued on next page}}} \\
\endfoot

    \bottomrule
\endlastfoot

    \citet{Nwoye.2020} & Laparoscopy & Triplet Recognition & 2D Dyn. \\
    \multicolumn{4}{p{\dimexpr\textwidth-2\tabcolsep}}{\textit{\footnotesize\textbf{Methods}: Introduces Triplet Rec. task; CNN (Tripnet) + 3D interaction space \quad \textbf{Datasets}: CholecT40}} \\
    \addlinespace[4pt]
    
    \citet{Islam.2020} & Robotics & Triplet Recognition & 2D Stat. \\
    \multicolumn{4}{p{\dimexpr\textwidth-2\tabcolsep}}{\textit{\footnotesize\textbf{Methods}: Graph-based reasoning + Feature Extraction Network (FEN) \quad \textbf{Datasets}: Custom annotated (EndoVis 2018)}} \\
    \addlinespace[4pt]

    \citet{Seenivasan.2022} & Robotics & Triplet Recognition & 2D Stat. \\
    \multicolumn{4}{p{\dimexpr\textwidth-2\tabcolsep}}{\textit{\footnotesize\textbf{Methods}: Graph-based Incremental Learning for Domain Adaptation \quad \textbf{Datasets}: Custom (Nephrectomy \& TORS)}} \\
    \addlinespace[4pt]
    
    \citet{Seenivasan.2022b} & Robotics & Triplet Recognition & 2D Stat. \\
    \multicolumn{4}{p{\dimexpr\textwidth-2\tabcolsep}}{\textit{\footnotesize\textbf{Methods}: Global-reasoned MTL (`GloRe`) for Rec. + Seg. \quad \textbf{Datasets}: MICCAI EndoVis 2018}} \\
    \addlinespace[4pt]

    \citet{Yang.2022} & GI Endoscopy & 3D Reconstruction & 2D Stat. \\
    \multicolumn{4}{p{\dimexpr\textwidth-2\tabcolsep}}{\textit{\footnotesize\textbf{Methods}: SG-driven feature matching for 3D Reconstruction \quad \textbf{Datasets}: Custom phantom \& clinical datasets}} \\
    \addlinespace[4pt]

    \citet{Nwoye.2022} & Laparoscopy & Triplet Recognition & 2D Stat. \\
    \multicolumn{4}{p{\dimexpr\textwidth-2\tabcolsep}}{\textit{\footnotesize\textbf{Methods}: Transformer-based Attn. (`Rendezvous`) \quad \textbf{Datasets}: CholecT50}} \\
    \addlinespace[4pt]

    \citet{Li.2022} & Laparoscopy & Triplet Recognition & 2D Stat. \\
    \multicolumn{4}{p{\dimexpr\textwidth-2\tabcolsep}}{\textit{\footnotesize\textbf{Methods}: Fine-grained interaction Rec. (`SIRNet`) \quad \textbf{Datasets}: CholecT50}} \\
    \addlinespace[4pt]

    \citet{Xi.2022} & Laparoscopy & Triplet Recognition & 2D Stat. \\
    \multicolumn{4}{p{\dimexpr\textwidth-2\tabcolsep}}{\textit{\footnotesize\textbf{Methods}: Classification Forest + GCN (`Forest GCN`) \quad \textbf{Datasets}: CholecT50}} \\
    \addlinespace[4pt]

    \citet{Pang.2022} & Laparoscopy & Triplet Recognition & 2D Stat. \\
    \multicolumn{4}{p{\dimexpr\textwidth-2\tabcolsep}}{\textit{\footnotesize\textbf{Methods}: Detector-free Rec. via gradient-based localization \quad \textbf{Datasets}: CholecT50}} \\
    \addlinespace[4pt]

    \citet{Lin.2022} & Robotics & Report Generation & 2D Dyn. \\
    \multicolumn{4}{p{\dimexpr\textwidth-2\tabcolsep}}{\textit{\footnotesize\textbf{Methods}: SG-guided Transformer (`SGT`) for report generation \quad \textbf{Datasets}: Custom (Nephrectomy from EndoVis 2018)}} \\
    \addlinespace[4pt]

    \citet{Holm.2023} & Microscopy & Workflow Recognition & 2D Dyn. \\
    \multicolumn{4}{p{\dimexpr\textwidth-2\tabcolsep}}{\textit{\footnotesize\textbf{Methods}: Dynamic SG + GCN for workflow analysis \quad \textbf{Datasets}: CaDIS \& CATARACTS}} \\
    \addlinespace[4pt]

    \citet{Wang.2023} & Robotics & Report Generation & 2D Dyn. \\
    \multicolumn{4}{p{\dimexpr\textwidth-2\tabcolsep}}{\textit{\footnotesize\textbf{Methods}: GNN-based relational exploration + Transformer \quad \textbf{Datasets}: 2018 MICCAI Robotic Instrument Seg. Challenge}} \\
    \addlinespace[4pt]

    \citet{Li.2023} & Laparoscopy & Triplet Recognition & 2D Dyn. \\
    \multicolumn{4}{p{\dimexpr\textwidth-2\tabcolsep}}{\textit{\footnotesize\textbf{Methods}: Multi-task framework (`MT-FIST`) with multi-label learning \quad \textbf{Datasets}: CholecT50}} \\
    \addlinespace[4pt]
    
    \citet{Seenivasan.2023} & Laparoscopy + Robotics & Triplet Recognition & 2D Dyn. \\
    \multicolumn{4}{p{\dimexpr\textwidth-2\tabcolsep}}{\textit{\footnotesize\textbf{Methods}: Asynchronous MTL + Incremental Contrastive Learning \quad \textbf{Datasets}: CholecT50, Custom robotic datasets}} \\
    \addlinespace[4pt]
    
    \citet{Sharma.2023} & Laparoscopy & Triplet Recognition & 2D Dyn. \\
    \multicolumn{4}{p{\dimexpr\textwidth-2\tabcolsep}}{\textit{\footnotesize\textbf{Methods}: Temporal fusion mechanism (`TAM`) to extend RDV model \quad \textbf{Datasets}: CholecT50}} \\
    \addlinespace[4pt]
    
    \citet{Sharma.2023b} & Laparoscopy & Triplet Recognition & 2D Dyn. \\
    \multicolumn{4}{p{\dimexpr\textwidth-2\tabcolsep}}{\textit{\footnotesize\textbf{Methods}: Two-stage Rec. (`R-CAG`) with pseudo-label generation \quad \textbf{Datasets}: CholecT50 \& a subset of Cholec80}} \\
    \addlinespace[4pt]
    
    \citet{Zou.2023} & Laparoscopy & Triplet Recognition & 2D Dyn. \\
    \multicolumn{4}{p{\dimexpr\textwidth-2\tabcolsep}}{\textit{\footnotesize\textbf{Methods}: End-to-end Spatio-Temporal Transformer \quad \textbf{Datasets}: CholecT45}} \\
    \addlinespace[4pt]
    
    \citet{Yuan.2024} & Laparoscopy & VQA & 2D Stat. \\
    \multicolumn{4}{p{\dimexpr\textwidth-2\tabcolsep}}{\textit{\footnotesize\textbf{Methods}: SG-based Interaction Module (`SIM`) for VQA \quad \textbf{Datasets}: SSG-VQA}} \\
    \addlinespace[4pt]

    \citet{Ban.2024} & Laparoscopy & Safety Assessment & 2D Dyn. \\
    \multicolumn{4}{p{\dimexpr\textwidth-2\tabcolsep}}{\textit{\footnotesize\textbf{Methods}: Concept Graph Network + Temporal concept integration \quad \textbf{Datasets}: CholecT45, CVS100, ParklandingGradeScale20}} \\
    \addlinespace[4pt]

    \citet{Ha.2024} & Laparoscopy & Triplet Recognition & 2D Dyn. \\
    \multicolumn{4}{p{\dimexpr\textwidth-2\tabcolsep}}{\textit{\footnotesize\textbf{Methods}: Model compression (Pruning + KD) of Rendezvous model \quad \textbf{Datasets}: CholecT50}} \\
    \addlinespace[4pt]

    \citet{Murali.2024} & Laparoscopy & Safety Assessment & 2D Stat. \\
    \multicolumn{4}{p{\dimexpr\textwidth-2\tabcolsep}}{\textit{\footnotesize\textbf{Methods}: Latent Graph Representation for Critical View of Safety (CVS) \quad \textbf{Datasets}: Endoscapes+ (extended for CVS)}} \\
    \addlinespace[4pt]

    \citet{Li.2024b} & Laparoscopy & Triplet Recognition & 2D Dyn. \\
    \multicolumn{4}{p{\dimexpr\textwidth-2\tabcolsep}}{\textit{\footnotesize\textbf{Methods}: Param-efficient framework (`LAM-Lite`) via lightweight CNN \quad \textbf{Datasets}: CholecT50}} \\
    \addlinespace[4pt]

    \citet{Gui.2024} & Laparoscopy & Triplet Recognition & 2D Stat. \\
    \multicolumn{4}{p{\dimexpr\textwidth-2\tabcolsep}}{\textit{\footnotesize\textbf{Methods}: Multi-Teacher Knowledge Distillation (`MT-FIST`) \quad \textbf{Datasets}: CholecT45}} \\
    \addlinespace[4pt]

    \citet{Liu.2024} & Laparoscopy & Triplet Recognition & 2D Dyn. \\
    \multicolumn{4}{p{\dimexpr\textwidth-2\tabcolsep}}{\textit{\footnotesize\textbf{Methods}: Generative framework via diffusion model (`DiffTriplet`) \quad \textbf{Datasets}: CholecT50}} \\
    \addlinespace[4pt]

    \citet{Koksal.2024} & Microscopy & Workflow Recognition & 2D Dyn. \\
    \multicolumn{4}{p{\dimexpr\textwidth-2\tabcolsep}}{\textit{\footnotesize\textbf{Methods}: Dynamic SG optimization with weak supervision \quad \textbf{Datasets}: CATARACTS}} \\
    \addlinespace[4pt]

    \citet{Lin.2024} & Robotics & Report Generation & 2D Dyn. \\
    \multicolumn{4}{p{\dimexpr\textwidth-2\tabcolsep}}{\textit{\footnotesize\textbf{Methods}: SG-guided Transformer (`SGT-SGG`) for report generation \quad \textbf{Datasets}: EndoVis-18 \& TORS}} \\
    \addlinespace[4pt]

    \citet{Zhang.2024b} & Laparoscopy & Triplet Recognition & 2D Dyn. \\
    \multicolumn{4}{p{\dimexpr\textwidth-2\tabcolsep}}{\textit{\footnotesize\textbf{Methods}: Spatio-temporal graph (`TAI-G`) for landmark tracking \quad \textbf{Datasets}: Custom Laparoscopic Cholecystectomy (LC) dataset}} \\
    \addlinespace[4pt]
    
    \citet{Lin.2024b} & Cataract & Triplet Recognition & 2D Dyn. \\
    \multicolumn{4}{p{\dimexpr\textwidth-2\tabcolsep}}{\textit{\footnotesize\textbf{Methods}: Quintuplet (instr-tissue-interaction) detection framework \quad \textbf{Datasets}: PhaCoQ \& CholecQ (new custom)}} \\
    \addlinespace[4pt]

    \citet{Wagner.2024} & Broad / General & Workflow Recognition & 2D Dyn. \\
    \multicolumn{4}{p{\dimexpr\textwidth-2\tabcolsep}}{\textit{\footnotesize\textbf{Methods}: Holistic multimodal graph for instrument anticipation \quad \textbf{Datasets}: Custom in-house cholecystectomy}} \\
    \addlinespace[4pt]

    \citet{Wang.2024} & Robotics & Ref. Inst. Segmentation & 2D Dyn. \\
    \multicolumn{4}{p{\dimexpr\textwidth-2\tabcolsep}}{\textit{\footnotesize\textbf{Methods}: \textit{VIS-Net}: Graph-based Relation-aware Module (GRM) correlates text descriptions with visual features. \quad \textbf{Datasets}: EndoVis-RS17, EndoVis-RS18}} \\
    \addlinespace[4pt]
    
    \citet{Pei.2025} & Laparoscopy & Triplet Recognition & 2D Dyn. \\
    \multicolumn{4}{p{\dimexpr\textwidth-2\tabcolsep}}{\textit{\footnotesize\textbf{Methods}: Tissue-guided triplet detection (`ITG-Trip`) \quad \textbf{Datasets}: CholecT50 \& Cholec80}} \\
    \addlinespace[4pt]
    
    \citet{Frisch.2025} & Microscopy & Gen. AI (SG to Image) & 2D Dyn. \\
    \multicolumn{4}{p{\dimexpr\textwidth-2\tabcolsep}}{\textit{\footnotesize\textbf{Methods}: Controllable SG-to-image diffusion model \quad \textbf{Datasets}: CaDIS}} \\
    \addlinespace[4pt]

    \citet{Zhang.2025} & Laparoscopy & Safety Assessment & 2D Dyn. \\
    \multicolumn{4}{p{\dimexpr\textwidth-2\tabcolsep}}{\textit{\footnotesize\textbf{Methods}: Knowledge-driven spatio-temporal graph (`TAI-G`) \quad \textbf{Datasets}: Custom Laparoscopic Cholecystectomy (LC) dataset}} \\
    \addlinespace[4pt]

    \citet{Yeganeh.2025} & Laparoscopy & Gen. AI (Video Synth.) & 2D Dyn. \\
    \multicolumn{4}{p{\dimexpr\textwidth-2\tabcolsep}}{\textit{\footnotesize\textbf{Methods}: Controllable video synthesis via action graph (`VISAGE`) \quad \textbf{Datasets}: CholecT50}} \\
    \addlinespace[4pt]

    \citet{Bai.2025} & Robotics & Workflow Recognition & 2D Dyn. \\
    \multicolumn{4}{p{\dimexpr\textwidth-2\tabcolsep}}{\textit{\footnotesize\textbf{Methods}: Multimodal graph learning for robust workflow Rec. \quad \textbf{Datasets}: MISAW \& CUHK-MRG}} \\
    \addlinespace[4pt]

    \citet{Hao.2025} & Robotics & VQA & 2D Stat. \\
    \multicolumn{4}{p{\dimexpr\textwidth-2\tabcolsep}}{\textit{\footnotesize\textbf{Methods}: Knowledge graph-guided LLM for specialized VQA \quad \textbf{Datasets}: EndoVis-17-VQLA \& EndoVis-18-VQLA}} \\
    \addlinespace[4pt]

    \citet{Shen.2025} & Robotics & Triplet Recognition & 2D Dyn. \\
    \multicolumn{4}{p{\dimexpr\textwidth-2\tabcolsep}}{\textit{\footnotesize\textbf{Methods}: Multiview integration network (`MI-Net`) for multimodal analysis \quad \textbf{Datasets}: EndoVis2018 \& private clinical dataset}} \\
    \addlinespace[4pt]

    \citet{Holm.2025} & Microscopy & Workflow Recognition & 2D Dyn. \\
    \multicolumn{4}{p{\dimexpr\textwidth-2\tabcolsep}}{\textit{\footnotesize\textbf{Methods}: GNN encoder processes dynamic SG; Prototype learning (clustering) in latent space. \quad \textbf{Datasets}: CAT-SG}} \\
    \addlinespace[4pt]

    \citet{Sivakumar.2025} & Cataract + Lap. & Generative AI & 2D Dyn. \\
    \multicolumn{4}{p{\dimexpr\textwidth-2\tabcolsep}}{\textit{\footnotesize\textbf{Methods}: \textit{SG2VID}: GNN-encoded SGs condition a Video Diffusion Model (3D U-Net) for synthesis. \quad \textbf{Datasets}: Cataract-1k, CATARACTS, Cholec80}} \\
    \addlinespace[4pt]

    \citet{Shin.2025} & Laparoscopy & Triplet Recognition & 2D Stat. \\
    \multicolumn{4}{p{\dimexpr\textwidth-2\tabcolsep}}{\textit{\footnotesize\textbf{Methods}: \textit{SSG-Com}: Detector identifies nodes; GCN predicts edges (spatial/action) and hand identity. \quad \textbf{Datasets}: Endoscapes-SG201}} \\
    \addlinespace[4pt]

    \citet{Biagini.2025} & Laparoscopy & Generative AI & 2D Dyn. \\
    \multicolumn{4}{p{\dimexpr\textwidth-2\tabcolsep}}{\textit{\footnotesize\textbf{Methods}: \textit{HieraSurg}: Hierarchical diffusion transformer (DiT) conditioned on encoded phases and triplets. \quad \textbf{Datasets}: Cholec80, CholecT45}} \\
    \addlinespace[4pt]

\end{longtable}

\begin{longtable}{@{} p{0.25\textwidth} p{0.2\textwidth} p{0.25\textwidth} p{0.12\textwidth} @{}}
    \caption{\textbf{Surgical Scene Graphs - External View Applications.}}
    \label{tab:summary_external_split} \\
    \toprule
    \textbf{Author \& Year} & \textbf{Domain} & \textbf{Application} & \textbf{SG Char.} \\
    \midrule
\endfirsthead

    \multicolumn{4}{c}%
    {{\tablename\ \thetable{} -- \textit{Continued from previous page}}} \\
    \toprule
    \textbf{Author \& Year} & \textbf{Domain} & \textbf{Application} & \textbf{SG Char.} \\
    \midrule
\endhead

    \bottomrule
    \multicolumn{4}{r}{{\textit{Continued on next page}}} \\
\endfoot

    \bottomrule
\endlastfoot

    \citet{Ozsoy.2021} & Broad / General & Scene Graph Generation & 4D Dyn. \\
    \multicolumn{4}{p{\dimexpr\textwidth-2\tabcolsep}}{\textit{\footnotesize\textbf{Methods}: MSSG Concept: Fuses visual, spatio-temporal, symbolic data \quad \textbf{Datasets}: MVOR}} \\
    \addlinespace[4pt]

    \citet{Ozsoy.2022} & Sim. Orthopedic & Scene Graph Generation & 4D Dyn. \\
    \multicolumn{4}{p{\dimexpr\textwidth-2\tabcolsep}}{\textit{\footnotesize\textbf{Methods}: Introduces 4D-OR dataset; Baseline SGG pipeline \quad \textbf{Datasets}: 4D-OR}} \\
    \addlinespace[4pt]
    
    \citet{Dsouza.2023} & Broad / General & Scene Graph Generation & 2D Stat. \\
    \multicolumn{4}{p{\dimexpr\textwidth-2\tabcolsep}}{\textit{\footnotesize\textbf{Methods}: Knowledge-graph integration for scene understanding \quad \textbf{Datasets}: Custom dataset of hospital scenes}} \\
    \addlinespace[4pt]
    
    \citet{Ozsoy.2023} & Sim. Orthopedic & Scene Graph Generation & 4D Dyn. \\
    \multicolumn{4}{p{\dimexpr\textwidth-2\tabcolsep}}{\textit{\footnotesize\textbf{Methods}: Memory-based SG for dynamic reasoning (`LABRAD-OR`) \quad \textbf{Datasets}: 4D-OR}} \\
    \addlinespace[4pt]

    \citet{Ozsoy.2024} & Sim. Orthopedic & Scene Graph Generation & 4D Dyn. \\
    \multicolumn{4}{p{\dimexpr\textwidth-2\tabcolsep}}{\textit{\footnotesize\textbf{Methods}: SGG pipeline from RGB-D for holistic OR modeling \quad \textbf{Datasets}: 4D-OR}} \\
    \addlinespace[4pt]

    \citet{Ozsoy.2024b} & Sim. Orthopedic & Scene Graph Generation & 4D Dyn. \\
    \multicolumn{4}{p{\dimexpr\textwidth-2\tabcolsep}}{\textit{\footnotesize\textbf{Methods}: LVLM for holistic guidance (`ORacle`) \quad \textbf{Datasets}: 4D-OR}} \\
    \addlinespace[4pt]

    \citet{Guo.2024} & Sim. Orthopedic & Scene Graph Generation & 4D Dyn. \\
    \multicolumn{4}{p{\dimexpr\textwidth-2\tabcolsep}}{\textit{\footnotesize\textbf{Methods}: Tri-modal confidence framework (`LLaVA-Med`) \quad \textbf{Datasets}: 4D-OR}} \\
    \addlinespace[4pt]

    \citet{Pei.2024} & Sim. Orthopedic & Scene Graph Generation & 4D Dyn. \\
    \multicolumn{4}{p{\dimexpr\textwidth-2\tabcolsep}}{\textit{\footnotesize\textbf{Methods}: Single-stage bi-modal Transformer (`S2Former-OR`) \quad \textbf{Datasets}: 4D-OR}} \\
    \addlinespace[4pt]
    
    \citet{Ozsoy.2025} & Sim. Robotic Orthopedic & Scene Graph Generation & 4D Dyn. \\
    \multicolumn{4}{p{\dimexpr\textwidth-2\tabcolsep}}{\textit{\footnotesize\textbf{Methods}: Introduces MM-OR dataset; Multimodal VLM baseline \quad \textbf{Datasets}: MM-OR (a new dataset)}} \\
    \addlinespace[4pt]
    
    \citet{Ozsoy.2025b} & Sim. Orthopedic & Foundation Models & 4D Dyn. \\
    \multicolumn{4}{p{\dimexpr\textwidth-2\tabcolsep}}{\textit{\footnotesize\textbf{Methods}: Foundation model for zero-shot performance (`OR-SOTA`) \quad \textbf{Datasets}: MM-OR, 4D-OR, EgoSurgery, MVOR}} \\
    \addlinespace[4pt]

    \citet{Ozsoy.2025c} & Sim. Orthopedic & Scene Graph Generation & 4D Dyn. \\
    \multicolumn{4}{p{\dimexpr\textwidth-2\tabcolsep}}{\textit{\footnotesize\textbf{Methods}: \textit{EgoExOR}: Dual-branch architecture fuses Ego+Exo features into LLM for SG generation. \quad \textbf{Datasets}: EgoExOR}} \\
    \addlinespace[4pt]

\end{longtable}
\endgroup

\par
\vspace{2mm}
\begin{center}
\footnotesize
    \textbf{Abbreviations:} 2D->3D (2D to 3D Recon. assist.), 2D Dyn. (2D Dynamic), 2D Imp. (2D Implicit), 2D Lat. (2D Latent), 2D Stat. (2D Static), 3D BBox (3D Bounding Box), 3D Recon. (3D Reconstruction), 3D Stat. (3D Static), 3D+T (3D + Time), 4D Dyn. (4D Dynamic), Annots (Annotations), ASR (Automatic Speech Recognition), Assist. (Assistance), Attn (Attention), Auto-Construct (Automatic Construction), BBox (Bounding Box), Cholec. (Cholecystectomy), CNN (Convolutional Neural Network), CT (Computed Tomography), CVS (Critical View of Safety), CXR (Chest X-Ray), Det. (Detection), GCN (Graph Convolutional Network), Gen. (General), GI (Gastrointestinal), GNN (Graph Neural Network), ICH (Intracranial Hemorrhage), Int. (Integration), KD (Knowledge Distillation), Lap. (Laparoscopic), LLM (Large Language Model), LSTM (Long Short-Term Memory), LVLM (Large Vision-Language Model), Mic. (Microscopic), MIS (Minimally Invasive Surgery), Mods (Modifications), MSSG (Multimodal Semantic Scene Graph), MTL (Multi-Task Learning), N/A (Not Applicable), Nephr. (Nephrectomy), OR (Operating Room), Orthopedic (Orthopedic Surgery), QA (Question Answering), Recog. (Recognition), Recon. (Reconstruction), Rel. (Relationships), Ref(s) (Reference(s)), Rob. (Robotic), Seg. (Segmentation), SG (Scene Graph), SGG (Scene Graph Generation), SIFT (Scale-Invariant Feature Transform), Sim. (Simulated), Spatio-Temporal (Space and Time), Stat./Dyn. (Static or Dynamic), Sup. (Supervised), TORS (Transoral Robotic Surgery), VLP (Vision-Language Pretraining), VQA (Visual Question Answering), VL (Vision-Language).
\end{center}

\bibliographystyle{elsarticle-harv} 
\bibliography{references}





\end{document}